\documentclass{article}
\usepackage{arxiv}
\usepackage{amsmath}
\usepackage[utf8]{inputenc} % allow utf-8 input
\usepackage[T1]{fontenc}    % use 8-bit T1 fonts
\usepackage{hyperref}       % hyperlinks
\usepackage{url}            % simple URL typesetting
\usepackage{booktabs}       % professional-quality tables
\usepackage{amsfonts}       % blackboard math symbols
\usepackage{nicefrac}       % compact symbols for 1/2, etc.
\usepackage{microtype}      % microtypography
\usepackage{lipsum}
\usepackage{graphicx}
\usepackage{algorithm}% http://ctan.org/pkg/algorithms
\usepackage{algpseudocode}% http://ctan.org/pkg/algorithmicx
\usepackage{float}
\usepackage{textcase}
\graphicspath{ {./images/} }

\title {Optimal Input Gain: All You Need to Supercharge a Feed-Forward Neural Network}
\author{
Chinmay Rane \\
Quantiphi Inc \\ 
Marlborough, MA, 01752 \\
\texttt{chinmay.rane@mavs.uta.edu}\\
   \And
Kanishka Tyagi,\\
  Aptiv Advance Research Center\\
  Agoura Hills, California, 91301\\
  \texttt{kanishka.tyagi@mavs.uta.edu} \\
  \And
Sanjeev Malalur, \\ 
Probe Technologies \\
Fort Worth, TX, 76140 \\
\texttt{sanjeev.malalur@gmail.com}\\
  \And
Yash Shinge \\
Amazon \\
Palo Alto, CA, 94020 \\ 
\texttt{yashpramod.shinge@mavs.uta.edu}\\
  \And
Michael Manry \\
Department of Electrical Engineering\\
The University of Texas at Arlington \\
Arlington, TX, 76010 \\
\texttt{manry@uta.edu}\\
}

\begin{document}

\maketitle
\begin{abstract}
Linear transformation of the inputs alters the training performance of feed-forward networks that are otherwise equivalent. However, most linear transforms are viewed as a pre-processing operation separate from the actual training. Starting from equivalent networks, it is shown that pre-processing inputs using linear transformation are equivalent to multiplying the negative gradient matrix with an autocorrelation matrix per training iteration. Second order method is proposed to find the autocorrelation matrix that maximizes learning in a given iteration. When the autocorrelation matrix is diagonal, the method optimizes input gains. This optimal input gain (OIG) approach is used to improve two first-order two-stage training algorithms, namely back-propagation (BP) and hidden weight optimization (HWO), which alternately update the input weights and solve linear equations for output weights. Results show that the proposed OIG approach greatly enhances the performance of the first-order algorithms, often allowing them to rival the popular Levenberg-Marquardt approach with far less computation. It is shown that HWO is equivalent to BP with Whitening transformation applied to the inputs. HWO effectively combines Whitening transformation with learning. Thus, OIG improved HWO could be a significant building block to more complex deep learning architectures.
\end{abstract}


\section{Introduction}\label{sec1}

Multilayer perceptron (MLP) neural networks are used to solve a variety of real-life approximation tasks, including stock market time series forecasting\cite{stock}, power load forecasting\cite{power_load_forecasting1}, prognostics\ \cite{prognostics}, well log processing \cite{well_logging2}, currency exchange rate prediction\cite{currency_ER}, control applications\cite{LEWIS_robotics} and stock and weather prediction\cite{forecasting}\cite{forecasting1}. MLPs are also used in classification problems such as speech recognition\cite{speech}, fingerprint recognition\cite{fingerprint}, character recognition\cite{character_recog}, and face detection\cite{face_recog}. In recent times, they also form the front end of deep learning architectures as \cite{tyagi2018automated}, \cite{qi2017pointnet}. 

The "no free lunch" theorem (NFL) \cite{duda2012pattern},\cite{wolpert1996lack} implies that no single neural network training algorithm is universally superior. Despite this, feed-forward neural nets, or MLPs, have gained increasing popularity for two reasons. First, MLPs have the ability to approximate any continuous function with arbitrary accuracy due to their universal approximation capability \cite{girosi1989representation}, \cite{Universal_Approx_theorem}, \cite{white1990connectionist}, \cite{hartman1990layered} meaning that it can approximate the best classifier. However, a feed-forward network with a single layer may struggle to learn and generalize correctly due to insufficient learning, a lack of deterministic relationship between inputs and outputs, or an insufficient number of hidden units \cite{pao1989adaptive}, \cite{werbos1974beyond}, \cite{GoodBengCour16}. Second, with proper training, an MLP can approximate the Bayes discriminant \cite{Bayes_Classifier} or the minimum mean-square error (MMSE) estimator \cite{geman1992neural},\cite{MSE1},\cite{MSE2}.

Several existing algorithms ranging from first order to second order have been developed for training MLPs, including output weight optimization-backpropagation (OWO-BP) \cite{hecht1992theory}, \cite{TYAGI20223}, scaled conjugate gradient (SCG) \cite{CG}\cite{moller1993scaled}, Levenberg-Marquardt (LM) \cite{LM1}\cite{LM2} and Newton's algorithm \cite{tyagi2021multistage}, \cite{tyagi2021multistage}. However, each of these approaches has its limitations. OWO-BP needs to scale better, i.e., it takes $O(N_w^2)$ operations for sufficiently large $N_w$, where $N_w$ is the total weights of the network. It is also unduly slow in a flat error surface and could be a more reliable learning paradigm. OWO-BP also lacks affine invariance \cite{TYAGI20223}. SCG scales well but has no internal mechanism to overcome linearly dependent inputs \cite{tyagi2018automated}. Newton's method is a second-order algorithm that requires computing and storing the Hessian matrix at each iteration, which can be computationally expensive \cite{Tan_2019}. The LM algorithm is faster than Newton's method because it approximates the Hessian matrix using the Jacobian matrix \cite{Levenberg1944}. However, it becomes less efficient as the number of parameters increases \cite{Tan_2019}. 

MLP training is also sensitive to many parameters of the network and its training data, including the input means \cite{Input_means}, the initial network weights \cite{BP1}, \cite{Pollack_Kolen}, and sensitive to the collinearity of its inputs \cite{Hashem1995ImprovingMA}. Also, scaling the network architecture to learn more complex representations is cumbersome. This limited applications involving MLP to challenging but less complex problems where shallow architectures could be used to learn and model the behavior effectively. The recent developments of transformers \cite{vaswani1601kaiser} and their success in complex applications involving natural speech \cite{dong2018speech}, \cite{pham2019very} and vision \cite{dosovitskiy2020image} have renewed interests in feed-forward network architectures, as they form the building blocks to the more complex transformer architectures \cite{introducingchatgpt}. Feed-forward network are also being used in active production for radar perception stack in autonomous driving. \cite{tyagi2022radar}. 

We present a family of fast learning algorithms targeted towards training a fixed architecture, fully connected multi-layer perceptron with a single hidden layer capable of learning from both approximation and classification datasets. In \cite{Optimal_Input_Gains}, a method for optimizing input gains called the optimal input gain (OIG) was presented. Preliminary experiments showed that when this method was applied to the first-order MLP training method like the back-propagation (BP), it significantly improved the overall network's performance with minimal computational overhead. However, the method performs less optimally under the presence of linearly dependent inputs. In general, this is true for other MLP training algorithms as well. We expand the idea of OIG to apply another first-order two-stage training algorithm called hidden weight optimization (HWO) \cite{HWO1}, \cite{TYAGI20223} to formulate the OIG-HWO algorithm. 

Following \cite{Optimal_Input_Gains}, we expand on the details of the motivation behind the OIG algorithm, along with a thorough analysis of its structure, performance, and limitation. In addition, we propose an improvement to overcome the limitation and compare the new algorithm's performance with existing algorithms. Our vision for the OIG-HWO presented in this paper is twofold, firstly, to be a strong candidate for challenging but less complex applications that can rival available shallow learning architectures in speed and performance, and secondly, to serve as a potential building block for more complex deep learning architectures.

The rest of the paper is organized as follows. Section I provides an introduction to the paper, while Section II covers the basics of MLP notation and training and an overview of existing algorithms. Section III discusses the linear transformation of inputs. In Section IV, we describe the OIG-HWO algorithm and its training process. Finally, in Section V, we compare the results of the OIG-HWO algorithm to those obtained using existing approaches on approximation data and classification data used for replacement in deep learning classifiers.

\section{Prior work}

\subsection{Structure and notation}

\begin{figure}[h!]
\begin{center}
\includegraphics[scale=0.4]{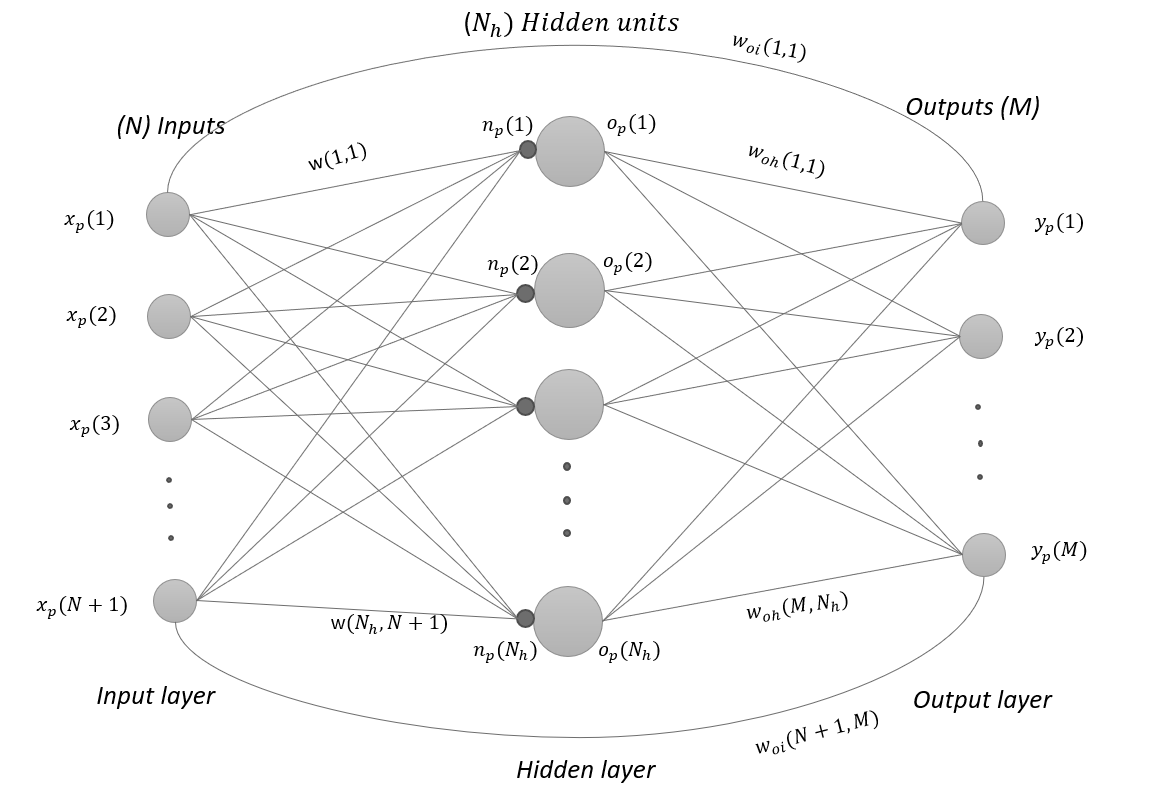}
\caption{Single hidden layer fully connected MLP}
\label{fig:Single layer MLP}
\end{center}
\end{figure}

A fully connected MLP with one hidden layer is shown in Figure \ref{fig:Single layer MLP}. Input weight $w(k, n)$ connects the $n^{th}$ input $x_p(n)$ to the $k^{th}$ hidden unit. Output weight $w_{oh}(i, k)$ connects the $k^{th}$ hidden unit’s activation $o_{p}(k)$ to the $i^{th}$ output $y_{p}(i)$, which has a linear activation. The bypass weight $w_{oi}(i, n)$ connects $x_p(n)$ to $y_p(i)$.
In the training data \{{$\mathbf{x}_{p}$, $\mathbf{t}_{p}$}\} for a fully connected MLP, the $p^{th}$ input vector $\mathbf{x}_{p}$ is initially of dimension N and the $p^{th}$ desired output (target) vector $\mathbf{t}_{p}$ has dimension M. The pattern number p varies from 1 to $N_v$. Let the input vectors be augmented by an extra element $x_{p}(N+1)$ where $x_{p}(N+1)$ = 1, so $\mathbf{x}_{p}$ = [$x_{p}(1)$, $x_{p}(2)$… $x_{p}(N+1)]^T$. Weights leading away from $x_{p}(N+1)$ are hidden or output layer thresholds. For the $p^{th}$ pattern, the hidden unit’s output vector $\mathbf{n_{p}}$ can be written as 

\begin{equation}
\begin{aligned}
\mathbf{n}_p = \mathbf{W} \cdot \mathbf{x}_p
\end{aligned}
\end{equation}

\noindent
where $\mathbf{n}_{p}$ is of size $N_{h}$ by $1$, and the input weight matrix \textbf{W} is of size $N_{h}$ by $(N+1)$. For the $p^{th}$ pattern, the $k^{th}$ hidden unit’s output, $o_{p}(k)$, is calculated as $o_{p}(k)$ = $f(n_{p}(k))$, where $f(.)$ denotes a nonlinear hidden layer activation function such as the rectified linear unit $(ReLU)$ which is given as follows \cite{Relu}.
\begin{equation}
\begin{aligned}
  f(n_p(k)) = max(0,n_p(k))=\begin{cases}
               n_p(k), \quad if \quad  n_p(k) \geq 0\\
               0, \quad if \quad n_p(k) <0\\
            \end{cases}
\end{aligned}
\end{equation}

\noindent
The $M$ dimensional network output vector $\mathbf{y_{p}}$ is 

\begin{equation}\label{eq:output_eq1}
\begin{aligned}
\mathbf{y}_p = \mathbf{W}_{oi} \cdot \mathbf{x}_p + \mathbf{W}_{oh} \cdot \mathbf{o}_p
\end{aligned}
\end{equation}

\noindent
where $\mathbf{o}_p$ is the $N_{h}$ dimensional hidden unit activation vector. The last columns of $\mathbf{W}$ and $\mathbf{W}_{oi}$ respectively store the hidden unit and output unit threshold values. During training the unknown weights are solved by minimizing a mean-squared error (MSE) function described as 
\begin{equation}\label{eq:MSE}
E = \frac{1}{N_v} \sum_{p=1}^{N_v}\sum_{i=1}^{M} [t_p(i) - y_p(i)]^2
\end{equation}

Training a neural network involves formulating it as an unconstrained optimization problem and then applying a learning procedure. Typically, the learning procedure is a line search \cite{gill2019practical}, with a layer-by-layer optimization \cite{biegler1993learning}, \cite{zhang1999learning}, \cite{wang1996fast}, involving first and second-order algorithms.

\subsection{Scaled conjugate gradient algorithm}
Conjugate gradient (CG) \cite{TYAGI20223} line-searches in successive conjugate directions and has faster convergence than steepest descent. To train an MLP using the CG algorithm (CG-MLP), we update all the network weights $\mathbf{w}$ simultaneously as follows: 
\begin{equation}
\mathbf{w} \leftarrow \mathbf{w} + z\cdot \mathbf{p} \\
\label{eq:cg-weightupdate}
\end{equation}
where $z$ is the learning rate that can be derived as \cite{Input_means},\cite{TYAGI20223}.
\begin{equation}
	\large
	z  = -{\frac{\frac{\partial E(\mathbf{w} + z \cdot \mathbf{p})}{\partial z}}{\frac{\partial^2 E(\mathbf{w} + z \cdot \mathbf{p})}{\partial z^2}}}\vert_{z=0}
	\label{eq:olf}
\end{equation}

\noindent
The direction vector $\mathbf{p}$ is obtained from the gradient $\mathbf{g}$ as 
\begin{equation}
	\mathbf{p} \leftarrow -\mathbf{g}  + B_1 \cdot \mathbf{p}
\end{equation}

\noindent
where $\mathbf{p}$ = $vec$ $(\mathbf{P},\mathbf{P}_\textrm{oh},\mathbf{P}_\textrm{oi})$ and $\mathbf{P}$, $\mathbf{P}_\textrm{oh}$ and $\mathbf{P}_\textrm{oi}$ are the direction vectors corresponding to weight arrays $(\mathbf{W},\mathbf{W}_\textrm{oh},\mathbf{W}_\textrm{oi})$. CG uses backpropagation to calculate $\mathbf{g}$. $B_1$ is the ratio of the gradient energy from two consecutive iterations. If the error function were quadratic, CG would converge in $N_w$ iterations \cite{Convex_opt}, where the number of network weights is $N_w$ = $dim(\mathbf{w})$. CG is scalable and widely used in training large datasets, as the network Hessian is not calculated \cite{Opt_DL_Andrew}. Therefore, in a CG, the step size is determined using a line search along the direction of the conjugate gradient. 

SCG \cite{moller1993scaled} scales the conjugate gradient direction by a scaling factor determined using a quasi-Newton approximation of the Hessian matrix. This scaling factor helps to accelerate the algorithm's convergence, especially for problems where the condition number of the Hessian matrix is large. SCG requires the computation of the Hessian matrix (or an approximation) and its inverse. A critical difference between CG and SCG is how the step size is determined during each iteration, with SCG using a scaling factor that helps to accelerate convergence. Other variations of CG exist \cite{tyagi2014optimal}. However, in this study, we choose to use SCG. 

\subsection{Levenberg-Marquardt algorithm}
The Levenberg-Marquardt (LM) algorithm \cite{TYAGI20223} is a hybrid first- and second-order training method that combines the fast convergence of the steepest descent method with the precise optimization of the Newton method \cite{Levenberg1944}. However, inverting the Hessian matrix $\mathbf{H}$ can be challenging due to its potential singularity or ill-conditioning \cite{Bishop_PRML}. To address this issue, the LM algorithm introduces a damping parameter $\lambda$ to the diagonal of the Hessian matrix as 
\begin{equation}
\mathbf{H}_{LM} = \mathbf{H} + \lambda \cdot \mathbf{I}
\end{equation}

\noindent
where $\mathbf{I}$ is an identity matrix with dimensions equal to those of $\mathbf{H}$. The resulting matrix $\mathbf{H}_{LM}$ is then nonsingular, and the direction vector $\mathbf{d}_{LM}$ can be calculated by solving:

\begin{equation}
\mathbf{H}_{LM} \mathbf{d}_{LM} = \mathbf{g}
\end{equation}

\noindent
The constant $\lambda$ represents a trade-off value between first and second order for the LM algorithm. When $\lambda$ is close to zero, LM approximates Newton's method and has minimal impact on the Hessian matrix. When $\lambda$ is large, LM approaches the steepest descent and the Hessian matrix approximates an identity matrix. However, the disadvantage of the LM algorithm is that it scales poorly and is only suitable for small data sets \cite{TYAGI20223}.

\subsection{Output weight optimization}
Output weight optimization (OWO) \cite{barton1991matrix}, \cite{tyagi2021multistage} is a technique to solve for $W_{oh}$ and $W_{oi}$ . Equation (\ref{eq:output_eq1}) can be re-written as
\begin{equation}
\begin{aligned}
\mathbf{y}_p = \mathbf{W}_o \cdot \mathbf{x}_{ap}
\end{aligned}
\end{equation}
where $\mathbf{x}_{ap}$ = $[\mathbf{x}_p^T : \mathbf{o}_p^T]^T$ is the augmented input column vector of size $N_u = N + N_h + 1$. 
$\mathbf{W}_o$ is formed as $[\mathbf{W}_{oi} : \mathbf{W}_{oh}]$ of dimensions $M$ by $N_u$. The output weights can be found by setting $\partial{E}/\partial{W_o} = 0$, which leads to a set of linear equations given by,
\begin{equation}\label{eq:C=RWo}
\begin{aligned}
\mathbf{C} = \mathbf{R} \cdot \mathbf{W}_o^T
\end{aligned}
\end{equation}
where  $\mathbf{C} = \frac{1}{N_v} \sum_{p=1}^{N_v} \mathbf{x}_{ap} \mathbf{t}_p^T$ and $\mathbf{R} = \frac{1}{N_v} \sum_{p=1}^{N_v} \mathbf{x}_{ap} \mathbf{x}_{ap}^T$. Equation (\ref{eq:C=RWo}) can be solved using orthogonal least squares (OLS) methods \cite{tyagi2018automated}. OWO provides fast training and avoids local minima \cite{Bayes_discriminant}. However, it only trains the output weights. 

\subsection{Input weight optimization}
Input weight optimization \cite{tyagi2018automated} is a technique for iteratively improving $\mathbf{W}$ via steepest descent. The $N_h$ by $(N +1)$ negative input weight Jacobian matrix
for the $p^{th}$ pattern's input weights is
\begin{equation}\label{eq:G_Jacobian}
\begin{aligned}
\mathbf{G} = \frac{1}{N_v} \sum_{p=1}^{N_v} \mathbf{\delta}_p \mathbf{x}_p^T
\end{aligned}
\end{equation}
where 
$\boldsymbol{\delta_p}$ = $[\delta_p(1),\delta_p(2),....,\delta_p(N_h)]^T$ is the $N_h$ by 1 column vector of hidden unit delta functions \cite{Hinton_ErrorProp}. $\mathbf{W}$ is updated in a given iteration as
\begin{equation}\label{eq:Weight_update}
\begin{aligned}
\mathbf{W} = \mathbf{W} + z \cdot \mathbf{G}
\end{aligned}
\end{equation}
where $z$ is the learning factor. Combined with BP, we formulate OWO-BP \cite{TYAGI20223}, a two-stage training algorithm developed as an alternative to BP. In a given iteration of OWO-BP, we first find the weights, $W_{oh}$ and $W_{oi}$ and then separately train $\mathbf{W}$ using BP \cite{tyagi2018automated}. OWO-BP is attractive for several reasons. First, the training is faster since solving linear equations for output weights in a given iteration is faster than using a gradient method. Second, when OWO optimizes output weights for a given input weight matrix, some local minima are avoided. Third, the method exhibits improved training performance compared to using only BP to update all the weights in the network. It can be shown that OWO-BP converges \cite{TYAGI20223}, and it leads to the convergence of the weights to a critical point in weight space \cite{TYAGI20223}. This can be a global minimum, a local minimum, or a saddle point.

\subsection{Hidden weight optimization}
\label{subsection:HWO}
HWO \cite{scalero1992fast}, finds an improved gradient matrix $\mathbf{G_{hwo}}$ by solving the following linear equation 
\begin{equation}\label{eq:D=GR-1}
\begin{aligned}
\mathbf{G}_{hwo} \cdot \mathbf{R_i} = \mathbf{G} 
\end{aligned}
\end{equation}
where $\mathbf{R_i}$ is the input autocorrelation matrix as
\begin{equation}\label{eq:r_i}
\mathbf{R_i} = \frac{1}{N_v} \sum_{p=1}^{N_v} \mathbf{x}_p  \mathbf{x}_p^T    
\end{equation}

and $\mathbf{G}$ is the backpropagation negative gradient matrix \cite{Hinton_ErrorProp}. Equation (\ref{eq:D=GR-1}) can be rewritten as 
\begin{equation}\label{eq:D=GR-12}
\begin{aligned}
\mathbf{G}_{hwo} = \mathbf{G} \cdot \mathbf{R_i}^{-1}
\end{aligned}
\end{equation}
\noindent

where $ \mathbf{G}_{hwo} = \mathbf{G} \cdot \mathbf{A}^T \cdot \mathbf{A}$. Equation (\ref{eq:D=GR-1}) can be solved using OLS or matrix inversion using the singular value decomposition (SVD). $\mathbf{A}$ is the whitening transform matrix \cite{raudys2001statistical}, \cite{tyagi2020second}. It is shown in \cite{tyagi2018automated} that HWO is equivalent to applying a whitening transform to the training data to de-correlate it. $\mathbf{W}$ is now updated using $\mathbf{G_{hwo}}$ instead  of $\mathbf{G}$  as 
\begin{equation}\label{eq:oig_Weight_update}
\begin{aligned}
\mathbf{W} = \mathbf{W} + z \cdot \mathbf{G}_{hwo}
\end{aligned}
\end{equation}

\section{Proposed work}
The study of the effects of applying the equivalent networks theory to the augmented input vectors $\mathbf{x}_p$ is thoroughly discussed in \cite{Optimal_Input_Gains}. In this work, we build upon the concept presented in \cite{Optimal_Input_Gains}, \cite{nguyen2016partially} and examine the impact of transformed input gains on the training process in conjunction with HWO. 

\subsection{Mathematical background}
Consider a nonlinear network designated as $\textit{MLP-1}$ with inputs $\mathbf{x} \in \mathbb{R}^{N+1}$, where the restriction $\mathbf{x}_{N+1} = 1$ is imposed, and outputs $\mathbf{y} \in \mathbb{R}^{M}$. Another network, referred to as $\textit{MLP-2}$, has inputs $\mathbf{x'} = \mathbf{A} \cdot \mathbf{x}$ and outputs $\mathbf{y'} \in \mathbb{R}^M$. These two networks are considered strongly equivalent if, for all $\mathbf{x}_p \in \mathbb{R}^{N+1}$, we have $\mathbf{y'}_p = \mathbf{y}_p$. The network $\textit{MLP-1}$ is trained on the original input vectors $\mathbf{x}_p$, while $\textit{MLP-2}$ is trained using the transformed input vectors $\mathbf{x'}_p$ defined as

\begin{equation}\label{eq:inp_data_trans}
\begin{aligned}
\mathbf{x'}_p = \mathbf{A} \cdot \mathbf{x}_p
\end{aligned}
\end{equation}
where $\mathbf{A}$ is an $N^{'}$ by $(N+1)$ rectangular transformation matrix, for some $N^{'}$ $\geq$ $(N+1)$. We establish in \cite{Optimal_Input_Gains} that input weights for \textit{MLP-1} and \textit{MLP-2} are related as 
\begin{equation}\label{eq:W'A=W}
\begin{aligned}
\mathbf{W'} \cdot \mathbf{A} = \mathbf{W}
\end{aligned}
\end{equation}

The negative gradient matrix for training the input weights in $\textit{MLP-2}$ is given by $\mathbf{G'} = \mathbf{G} \cdot \mathbf{A}^T$. Now, suppose that this negative gradient $\mathbf{G'}$ for $\textit{MLP-2}$ is mapped back to modify the input weights in $\textit{MLP-1}$, using equation (\ref{eq:W'A=W}). The resulting mapped negative gradient for $\textit{MLP-1}$ is then

\begin{equation}\label{eq:G"=GR}
\begin{aligned}
\mathbf{G''} &= \mathbf{G} \cdot \mathbf{R}_i\\
\mathbf{R}_i &= \mathbf{A}^T \cdot \mathbf{A}
\end{aligned}
\end{equation}

By expressing the SVD of $\mathbf{R_i}$ as $\mathbf{R_i} = \mathbf{U \Sigma U}^T$, we can derive that $\mathbf{R_i}^{-1} = \mathbf{U \Sigma^{-1} U}^T$, where $\mathbf{U}$ is an orthogonal matrix, $\mathbf{\Sigma}$ is a diagonal matrix with the singular values of $\mathbf{R_i}$, and $\mathbf{\Sigma^{-1}}$ is the diagonal matrix with the reciprocal of the non-zero singular values of $\mathbf{R_i}$. Using equation (\ref{eq:G"=GR}), it can be deduced that $\mathbf{A} = \mathbf{\Sigma}^{\frac{1}{2}} \mathbf{U}^T$. Comparing equation (\ref{eq:G"=GR}) with equation (\ref{eq:D=GR-12}), it is clear that performing OWO-HWO is equivalent to performing OWO-BP on input vectors to which the whitening transformation \cite{tyagi2020second} has been applied. Since BP with optimal learning factor (OLF) converges, it is clear that HWO with an OLF also converges.
 
\textit{\textbf{Lemma-1}}:  \textit{If we are at a local minimum in the weight space of the original network, we are also at a local minimum in the weight space of the transformed network.}

This follows from ((\ref{eq:G"=GR}) if G = 0.

\textit{\textbf{Lemma-2}}: \textit{If the input weight matrix $\mathbf{W'}$ of the transformed network is trained using BP, this is not equivalent to applying BP to the original network's weight matrix $\mathbf{W}$ unless the matrix $\mathbf{A}$ is orthogonal}. 

This can be derived from equation (\ref{eq:G"=GR}) because for any orthogonal matrix $\mathbf{A}$, equation (\ref{eq:G"=GR}) becomes $\mathbf{G''} = \mathbf{G}$. This is also intuitive if we consider that BP updates the weights in the direction of the negative gradient of the loss function with respect to the weights. 

\textit{\textbf{Lemma-3}}: \textit{For a non-diagonal matrix $\mathbf{R}$, there exist an uncountably infinite number of matrices $\mathbf{A}$ that can be constructed.}

This follows from (\ref{eq:G"=GR}). It is because a non-diagonal matrix $\mathbf{R}$ has at least one non-zero element not located on the main diagonal. As there are infinite choices for the values of these non-zero elements, there are an uncountable number of possible matrices $\mathbf{A}$ that can be constructed by choosing the values of these non-zero elements in $\mathbf{R}$.

If the transformation matrix $\mathbf{A}$ is not orthogonal, then the mapped negative gradient for $\textit{MLP-1}$ obtained from $\textit{MLP-2}$ will not be equivalent to the true negative gradient of the loss function with respect to the weights in $\textit{MLP-1}$. As a result, when optimal learning factors are used with BP to train the input weights, training with the original data is equivalent to training with the transformed data. Therefore, orthogonal transform matrices are useless in this context, as mentioned in \cite{Manry_Effects_of_NonSing_FFN}. Using the results derived in this section, many ways to improve feed-forward training algorithms suggest themselves. The intuition behind the proposed work is based on the following three ideas : 
\begin{enumerate}
    \item Choose a training algorithm that utilizes the negative gradient matrix $\mathbf{G}$.
    \item Substitute the negative gradient matrix $\mathbf{G}$ with the modified matrix $\mathbf{G''}$ from equation (\ref{eq:G"=GR}).
    \item  Identify appropriate elements in the matrix $\mathbf{R}$.
\end{enumerate}   
 
Single-stage optimization algorithms, such as conjugate gradient (CG) \cite{TYAGI20223}, may be suitable for addressing this problem. However, incorporating the elements of $\mathbf{R}$ as additional weights in the optimization process may compromise the conjugacy of the direction vectors if $\mathbf{R}$ is solved for at each iteration. As an alternative, using two-stage training algorithms that utilize the negative gradient matrix $\mathbf{G}$ or direction matrix $\mathbf{D}$, such as OWO-BP  and OWO-HWO \cite{Manry_Multi_Linear}. In this work, we focus on OIG to develop OIG-HWO. Specifically, we will develop a method for solving the matrix $\mathbf{R}$, compute the resulting Gauss-Newton approximate Hessian for $\mathbf{R}$, and apply the resulting OIG-HWO to improve the performance of OWO-BP. 

\subsection{Optimal Input Gain algorithm}

There are at least two intuitive approaches for optimizing input gains to improve the performance of a given training algorithm. To minimize the training error $ E$, these approaches involve searching for either the matrix $\mathbf{A}$ or the resulting matrix $\mathbf{R}$ in each iteration to minimize the training error $E$. As stated in \textit{Lemma-2}, optimizing $\mathbf{R}$ will likely yield fewer solutions. In this section, we describe a method for solving $\textbf{R}$, find the resulting Gauss-Newton approximate Hessian for $\textbf{R}$, and use the resulting OIG algorithm to improve OWO-BP. The simplest non-orthogonal, non-singular transform matrix $\mathbf{A}$ is diagonal. For this case, let $r(k)$ initially denote the $k^{th}$ diagonal element of $\mathbf{R}$. Also, the elements of $\mathbf{x'}_p$ are simply scaled versions of $\mathbf{x}_p$. Following 

(\ref{eq:G"=GR}) we get
\begin{equation}
\begin{aligned}
\textbf{\textit{R}} =
  \begin{bmatrix}
    r(1) & 0 & \cdots & 0 & 0 \\
    0 & r(2) & \cdots & 0 & 0 \\
    \vdots & \vdots & \ddots & \vdots & \vdots \\
    0 & 0 & \cdots & r(N) & 0\\
    0 & 0 & \cdots & 0 & r(N+1)
  \end{bmatrix}
\end{aligned}
\end{equation}

Instead of using only the negative gradient elements $g(k, n)$ to update the input weights, we use $g(k, n) \cdot r(n)$ to replace $g(k, n)$, the elements matrix $\mathbf{G}$ in equation \ref{eq:oig_Weight_update}. It is also noteworthy that the optimal learning factor (OLF), $z$ \cite{TYAGI20223} be absorbed into the gains $r(n)$. Consider a multi-layer perceptron (MLP) being trained using the OWO-BP algorithm. The negative gradient $\mathbf{G}$ is a matrix of dimensions $N_h$ by $(N+1)$, and the error function to be minimized with respect to the gains $r(n)$ is given in (\ref{eq:MSE}). This error function is defined as follows:

\begin{equation}
\begin{aligned}
y_p(i) = \sum_{n=1}^{N+1}w_{oi}(i,n) x_p(n) + 
\sum_{k=1}^{N_h}w_{oh}(i,k)\cdot \\ 
f\left( \sum_{n=1}^{N+1}(w(k,n) + r(n) \cdot g(k,n)) x_p(n)\right)
\end{aligned}
\end{equation}

The first partial of $E$ with respect to $r(m)$ is
\begin{equation}\label{eq:dE_w.r.t_r}
\begin{aligned}
d_r(m) \equiv \dfrac{\partial{E}}{\partial{r(m)}} = \dfrac{-2}{N_v}\sum_{p=1}^{N_v} x_p(m) \\
\sum_{i=1}^{M} [t_p(i) - y_p(i)] v(i,m)
\end{aligned}
\end{equation}
\noindent
Here, $g(k, m)$ is an element of the negative gradient matrix \textbf{G} in equation (\ref{eq:G_Jacobian}), and $o^{'}_p(k)$ denotes the derivative of $o_p(k)$ with respect to its net function. Then, 
\begin{equation}
\begin{aligned}
v(i,m) = \sum_{k=1}^{N_h} w_{oh}(i,k) o^{'}_p(k) g(k,m)
\end{aligned}
\end{equation}
\noindent
Using Gauss-Newton updates \cite{Bishop_PRML}, the elements of the Hessian matrix $\bf{H_{ig}}$ are
\begin{equation}\label{eq:GN_Hess}
\begin{aligned}
h_{ig}(m,u) \equiv \dfrac{\partial^2{E}}{\partial{r(m)} \partial{r(u)} } = \dfrac{2}{N_v} \sum_{p=1}^{N_v} x_p(m) x_p(u) \cdot \\ \sum_{i=1}^{M} v(i,m) v(i,u)
\end{aligned}
\end{equation}

Finally, the input gain coefficient vector $\mathbf{r}$ is calcualted using OLS by solving

\begin{equation} \label{eq:oig_equation}
\begin{aligned}
    \mathbf{H_{ig}} \cdot  \mathbf{r}  = \mathbf{d}_r
\end{aligned}
\end{equation}

\subsubsection{OIG Hessian matrix}
We choose to use Hessian matrix to analyze the convergence properties of OIG-HWO. Equation (\ref{eq:GN_Hess}) for the OIG-HWO Hessian can be re-written as, 

\begin{equation}
\begin{aligned}
h_{ig}(m,u) = \sum_{k=1}^{N_h} \sum_{j=1}^{N_h} \left[ \dfrac{2}{N_v} \sum_{p=1}^{N_v} x_p(m) x_p(u) o_p^{'}(k) o_p^{'}(k)\cdot \sum_{i=1}^{N+1} w_{oh}(i,k) w_{oh}(i,j) \right] g(k,m) \cdot g(j, u)
\end{aligned}
\end{equation}
The term within the square brackets is nothing but an element from the Hessian of Newton’s method for updating input weights. Hence,
\begin{equation}\label{eq:4_ind_Newton_Hess}
\begin{aligned}
h_{ig}(m,u) = \sum_{k=1}^{N_h} \sum_{j=1}^{N_h} \left[
\dfrac{ \partial^2{E} }{ \partial{w(k,m)} \partial{w(j,u)} }
\right]\cdot g(k,m) \cdot g(j, u)
\end{aligned}
\end{equation}
\noindent

For fixed $(m, u)$, the above equation can be expressed as 
\begin{equation}\label{eq:OIG-HWO_Hess_1}
\begin{aligned}
h_{ig}(m,u) &= \sum_{k=1}^{N_h} g_m(k) \sum_{j=1}^{N_h}
h_N^{m,u}(k,j) g_u(j)\\
&= \mathbf{g_m^T H_N^{mu} g_u}
\end{aligned}
\end{equation}
\noindent
where, $g_m$ is the $m^{th}$ column of the negative gradient matrix $\mathbf{G}$ and $\mathbf{H}_N^{m,u}$ is the matrix formed by choosing elements from the Newton’s Hessian for weights connecting inputs $(m, u)$ to all hidden units.

Equation (\ref{eq:OIG-HWO_Hess_1}) gives the expression for a single element of the OIG-HWO Hessian, which combines information from $N_h$ rows and columns of the Newton Hessian. This can be seen as compressing the original Newton Hessian of dimensions $N_h$ by $(N+1)$ down to $(N+1)$. The OIG-HWO Hessian encodes the information from the Newton Hessian in a smaller dimension, making it less sensitive to input conditions and faster to compute. From equation (\ref{eq:4_ind_Newton_Hess}), we see that the Hessian from Newton’s method uses four indices $(j, m, u, k)$ and can be viewed as a 4-dimensional array, represented by $H_N^{4}$ $\in$ $\mathbf{R}^{N_h\text{x}(N+1)\text{x}(N+1)\text{x}N_h} $. Using this representation, we can express a 4-dimensional OIG-HWO Hessian as 

\begin{equation}
\begin{aligned}
\mathbf{H}^4_{ig} = \mathbf{G}^T \mathbf{H}_N^4 \mathbf{G}
\end{aligned}
\end{equation}
where $\mathbf{H}^4_{ig}$ are defined as,
\begin{equation}\label{eq:H4_ig_}
\begin{aligned}
h^4_{ig}(m,u,n,l) = \sum_{j=1}^{N_h} \sum_{k=1}^{N_h}
h_N(j,m,u,k) g(j,n) g(k,l)
\end{aligned}
\end{equation}
\noindent
where $\mathit{h_N(j, m, u, k)}$ is an element of $\mathbf{H}_N^4$. Comparing (\ref{eq:4_ind_Newton_Hess}) and (\ref{eq:H4_ig_}), we see that $h_{ig}(m, u)$ = $h^{4}_{ig}(m, u, m, u)$, i.e., the 4-dimensional $\mathbf{H}^4_{ig}$ is transformed into the 2-dimensional Hessian, $\mathbf{H}_{ig}$, by setting $n = m$ and $l = u$. To make this idea clear, consider a matrix, $\mathbf{Q}$, then $p(n) = q(n, n)$ is a vector, $\mathbf{p}$, of all diagonal elements of $\mathbf{Q}$. Similarly, the OIG-HWO Hessian $\mathbf{H}_{ig}$ is formed by a weighted combination of elements from $\mathbf{H}^4_{N}$.

\subsubsection{OIG Integrated with OWO-BP}
To minimize the error function $E$, given the vector of input gains $\mathbf{r}$, the gradient $\mathbf{d}_r$, and the Hessian $\mathbf{H}_{ig}$, we can utilize Newton's method. Two potential approaches can be taken in each iteration of this method, first is that we transform the gradient matrix using $\mathbf{R}$ as shown in equation (\ref{eq:G"=GR}), and second, we decompose $\mathbf{R}$ to find $\mathbf{A}$ using OLS, and then transform the input data according to equation (\ref{eq:inp_data_trans}) before applying OWO-BP with the optimal learning factor (OLF). While the second approach may be more accurate, it is also more computationally inefficient and, therefore, not practical, even when $\mathbf{A}$ is diagonal. Therefore, it is generally recommended to use the first approach in order to minimize the error function effectively. 
Hence, the OWO is replaced with OIG in the OWO-BP algorithm to form OIG-BP described in Algorithm \ref{algo:oig-bpalgorithm}. 

\begin{algorithm}[H]
	\caption{OIG-BP training algorithm}
	\label{algo:oig-bpalgorithm}
	\begin{algorithmic}[1]
		\State Initialize $\mathbf{W, W_{oi}, W_{oh}}$, $N_{it}$ , it$\leftarrow$ 0
		\While{it $<$ $N_{it}$}
        \State Solve (\ref{eq:C=RWo}) for all output weights. 
        \State Calculate negative $\mathbf{G}$ using equation (\ref{eq:G_Jacobian})
        \State $\textbf{OIG step}$ Calculate $\mathbf{d_r}$ and hessian $\mathbf{H}_{ig}$ from (\ref{eq:dE_w.r.t_r}) and (\ref{eq:GN_Hess}) respectively. 
        \State Solve for $\mathbf{r}$ using equation (\ref{eq:oig_equation})
		\State Update $\mathbf{W}$ $\leftarrow$ $\mathbf{W}$ + $\mathbf{r}$ $\cdot$ $\mathbf{G}$
		\State $\textbf{OWO step}$ : Solve equation (\ref{eq:C=RWo}) to obtain $\mathbf{W_o}$ 
		\State it $ \leftarrow $ it + 1 
		\EndWhile
	\end{algorithmic}
\end{algorithm}

When there are no linearly dependent inputs, the OIG algorithm can find the optimal gain coefficients for each input that minimize the overall mean squared training error. However, this is only sometimes the case when there are linearly dependent inputs. In this scenario, it is straightforward to show that the input autocorrelation matrix $\mathbf{R}_{in}$ and the gradient matrix $\mathbf{G}$ have dependent columns. This leads to the OIG Hessian being \textit{ill-conditioned} and to sub-optimal gain coefficients. This could cause OIG-BP to have sub-optimal performance and possibly poor convergence.

\subsection{Improvement to OIG-BP}
In order to overcome sub-optimal performance of OIG in the presence of linearly dependent inputs, we show the immunity of HWO to linearly dependent inputs. We analyze the effect of replacing BP used in OIG-BP with HWO and show that using HWO forces $H_{ig}$ to be singular for linearly dependent inputs, which is highly desirable in order to detect and eliminate the dependent inputs.

\subsubsection{Effect of Linearly Dependent Inputs on HWO}
If one of the inputs to the network is linearly dependent, it will cause the input auto-correlation matrix, $\mathbf{R}_{i}$, to be singular. This can affect the convergence of the CG algorithm, leading to poor training performance. In this case, using OLS may be useful for detecting and eliminating the linearly dependent input. To compute the orthonormal weight update matrix $\mathbf{G}_{hwo}$ using OLS, we first compute $\mathbf{G}_{hwo}^{'}$ as 

\begin{equation}
\mathbf{G}_{hwo}' = \mathbf{G} \cdot \mathbf{C}^T    
\end{equation}
where $\mathbf{C}$ is a lower triangular matrix of orthonormal coefficients of dimension $(N+1)$. We can then map the orthonormal weight update to the original weight update as 
\begin{equation} \label{eq:G_hwo}
\begin{aligned}
\mathbf{G}_{hwo} = \mathbf{G}_{hwo}' \cdot \mathbf{C}\\
=\mathbf{G} \cdot \mathbf{C}^T \cdot \mathbf{C}
\end{aligned}
\end{equation}
\noindent

Assume $\mathbf{x}_p(N+2)$ was linearly dependent. This would cause the $(N+2)^{th}$ row and column of $\mathbf{R}_{i}$ to be linearly dependent. During OLS, a singular auto-correlation matrix transforms to the $(N+2)^{th}$ row of $\mathbf{C}$ to be zero. We replace BP in OIG-BP with HWO. The resulting OIG-HWO algorithm is described in Algorithm \ref{algo:owo-hwoalgorithm}. 
\begin{algorithm}[H]
	\caption{OIG-HWO training algorithm}
	\label{algo:owo-hwoalgorithm}
	\begin{algorithmic}[1]
		\State Initialize $\mathbf{W, W_{oi}, W_{oh}}$, $N_{it}$ , it$\leftarrow$ 0
        \State Calculate $\mathbf{R}_i$ using (\ref{eq:r_i})
		\While{it $<$ $N_{it}$}
        \State Calculate negative $\mathbf{G}$ using (\ref{eq:G_Jacobian})
		\State $\textbf{HWO step}$ : Calculate $\mathbf{G_{hwo}}$ using  (\ref{eq:G_hwo}) to eliminate any linear dependency in the inputs.
        \State $\textbf{OIG step}$: Calculate $\mathbf{d_r}$ and hessian $\mathbf{H}_{ig}$ from (\ref{eq:dE_w.r.t_r}) and (\ref{eq:GN_Hess}) respectively. 
        \State Solve for $\mathbf{r}$ using equation (\ref{eq:oig_equation})
		\State Update $\mathbf{W}$ $\leftarrow$ $\mathbf{W}$ + $\mathbf{r}$ $\cdot$ $\mathbf{G_{hwo}}$
		\State $\textbf{OWO step}$ : Solve equation (\ref{eq:C=RWo}) to obtain $\mathbf{W_o}$ 
		\State it $ \leftarrow $ it + 1 
		\EndWhile
	\end{algorithmic}
\end{algorithm}

\textbf{\textit{Lemma-3}}: \textit{The $OIG-HWO$ algorithm is immune to linearly dependent inputs and will completely ignore the dependent inputs during training. }

Since the $(N+2)^{th}$ row of $\mathbf{C}$ will be zero, it follows that $\mathbf{C}^T\mathbf{C}$, which will be a square, symmetric matrix with zeros for the $(N+2)^{th}$ row and column. Further, from (\ref{eq:G_hwo}), $\mathbf{G}_{hwo}$ will have zeros for the $(N+2)^{th}$ column. The implication is that the weight update vector computed for all input weights connected to the dependent input $(N+2)$ is zero. These weights are not updated during training, effectively \textit{freezing} them. This is highly desirable, as the dependent input does not contribute any new information. Thus, HWO-type update using OLS is perfectly capable of picking up linearly dependent inputs, leading to a robust training algorithm. This makes OIG-HWO immune to linearly dependent inputs. 

\begin{figure}[H] 
\begin{center}
\centerline{\includegraphics[scale=1]{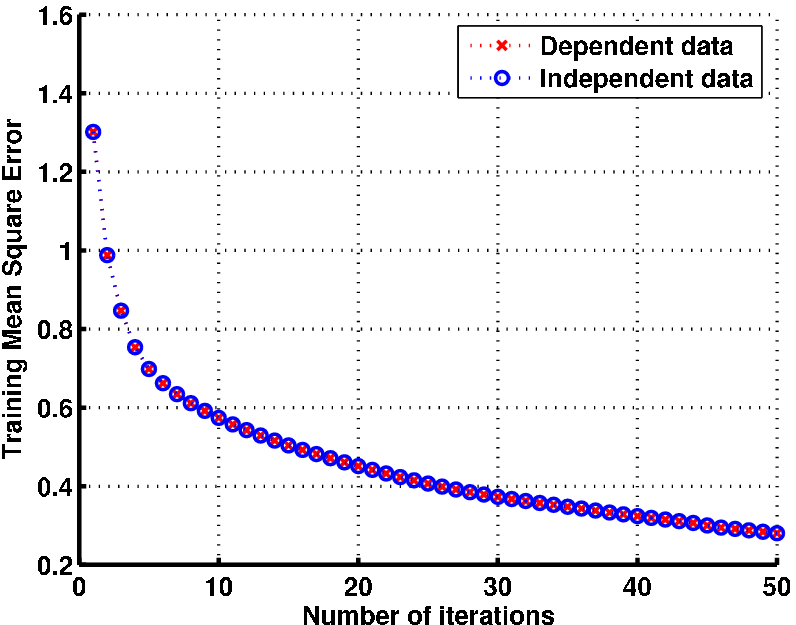}}
\caption{Immunity of OIG-HWO to linearly dependent inputs}
\label{fig:dependen1}
\end{center}
\end{figure}

To illustrate the meaning of \textit{lemma-3}, we took a data set called $\mathit{twod.tra}$ \cite{ipnnl_dataset_approx}, and generated a second one by adding some dependent inputs. Networks for the two datasets were initialized with the same net function means and standard deviations. Figure \ref{fig:dependen1} clearly shows that the two training error curves overlay each other, validating \textit{lemma-3}. 

\begin{figure}[H] 
\begin{center}
\centerline{\includegraphics[scale=1]{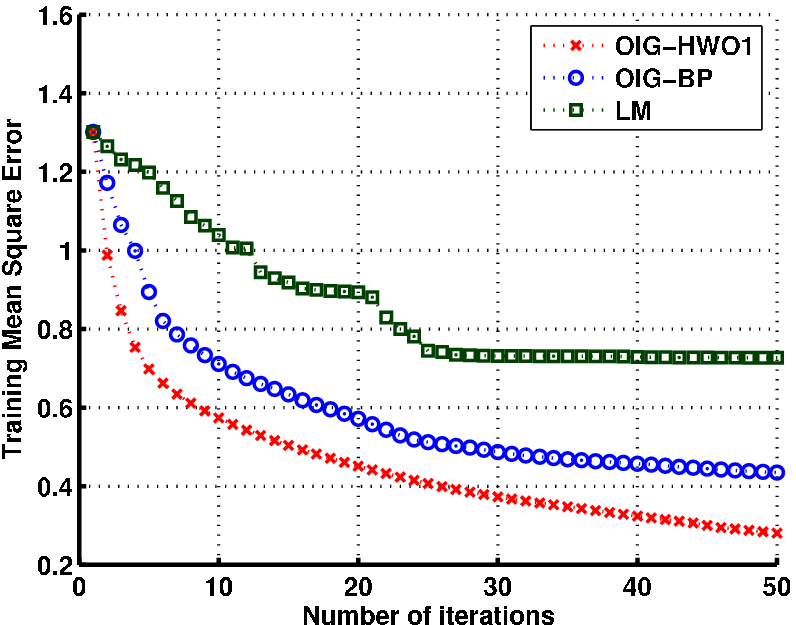}}
\caption{Performance comparison on dependent data}
 \label{fig:dependen2}
\end{center}
\end{figure}

To further demonstrate \textit{lemma-3} and the effectiveness of OIG-HWO, we compare its performance on the dependent dataset with LM and OIG-BP. Figure \ref{fig:dependen2} shows how dependence can slow down learning in all except the improved OIG-HWO algorithm. The effect is predominant in LM that takes huge computational resources. 

Mathematically, suppose that the input vector $\mathbf{x_p}$ are biased such that $E[\mathbf{x}_p]$ = $\mathbf{m}$. A zero-mean version of $\mathbf{x_p}$ is $\mathbf{x}'_p$ which satisfies  $\mathbf{x}_p$  =  $\mathbf{x}'_p$ +  $\mathbf{m}$.  It is shown in \cite{Optimal_Input_Gains} that networks train more effectively with bunbiased inputs. Now,  $\mathbf{x}'_p$  can be expressed as   $\mathbf{A}$  $\cdot$  $\mathbf{x}_p$, where
\begin{equation}
\begin{aligned}
\textbf{A} =
  \begin{bmatrix}
    1 & 0 & \cdots & 0 & -m_1 \\
    0 & 1 & \cdots & 0 & -m_2 \\
    \vdots & \vdots & \ddots & \vdots & \vdots \\
    0 & 0 & \cdots & 1  & -m_N\\
    0 & 0 & \cdots & 0 & 1
  \end{bmatrix}
\end{aligned}
\end{equation}

Figure \ref{fig:dependen2} shows that non-orthogonal transform matrices improve the training since it makes the inputs zero-mean. The HWO component of the OIG-HWO algorithm addresses the issue of sub-optimal performance in the presence of linearly dependent inputs. It has been demonstrated that the HWO is immune to such inputs \cite{Optimal_Input_Gains}. By replacing the BP component of the OIG-BP with HWO, we can analyze the effect on the singularity of $\mathbf{H}_{ig}$ for linearly dependent inputs. This is beneficial because the singularity of $\mathbf{H}_{ig}$ allows for the detection and removal of dependent inputs.

\section{Experimental Methods and Results}
\label{sec:Experimental Results}
We evaluate the computational complexity and performance of the OIG-HWO algorithm compared to other methods for approximation and replacement classifier tasks. In a replacement classifier, the OIG-HWO is used as a substitute in a ResNet18 \cite{resnet} style deep learning architecture and compares its testing performance to that of a scaled-CG (SCG) classifier \cite{tyagi2021multistage}. We compare the performance of the proposed OIG-HWO algorithm with four other existing methods, namely, OWO-BP, OIG-BP, LM, and SCG. In SCG and LM, all weights are updated simultaneously in each iteration. However, in OWO-BP, OIG-BP, and OIG-HWO, we solve linear equations for the output weights and then update the input weights.  

\subsection{Experimental procedure}
All the experiments are run on a machine equipped with a 3 MHz Intel i-7 CPU and 32 GB of RAM running the Windows 10 OS with PyTorch 1.13. We use the $\textit{k-fold}$ training with cross-validation and testing procedures to obtain the average training, validation, and testing errors. Each data set is split into $\textit{k}$ non-overlapping parts of equal size ($\textit{k}$ = 10 in our simulations). Of this, ($\textit{k}$-2) parts (roughly 80\%) are used for training. Of the remaining two parts, one is used for validation, and the other is used for testing (roughly 10\% each). The procedure is repeated till we have exhausted all $\textit{k}$ combinations. 

Validation is performed per training iteration (to prevent over-training), and the network with the minimum validation error is saved. After training, the saved weights and the testing data are used to compute a testing error, which measures the network's ability to generalize. At the end of the $\textit{k}$-fold procedure, the average training and testing errors and the average number of cumulative multiplies required for training are computed. These quantities form the metrics for comparison and are subsequently used to generate the plots and compare performances.

In order to make a fair comparison of the various training methods for a given data set and fold, we use the same initial network for each algorithm, using net control \cite{TYAGI20223}. In net control, random initial input weights and hidden unit thresholds are scaled so each hidden unit's net function has the same mean and standard deviation. Specifically, the net function means are equal to .5, and the corresponding standard deviations are equal to 1. In addition, OWO is used to find the initial output weights. This ensures that all algorithms start at the same point in the weight space, eliminating any performance gains due to weight initialization. This is evident in all the plots, where we can see that all algorithms have the same starting MSE for the first training iteration. The final training error is hence not affected by different weight initializations. For each approximation datasets, we calculate the lowest MSE and probability of error, $Pe$ for classification datasets.

\subsection{Computational Burden}
One of the metrics chosen for comparison is the cumulative number of multiplies required for training using a particular algorithm. In this section, we identify the computational burden per training iteration for each of the algorithms compared.

Let $N_u$ = $N+N_h+1$ denote the number of weights connected to each output. The total number of weights in the network is denoted as $N_w$ = $M(N+N_h+1) + N_h(N+1)$. The number of multiplies required to solve for output weights using OLS is $M_{ols}$, which is given by 
\begin{equation}
\begin{aligned}
M_{ols} = N_u (N_u + 1) \left[ M + \frac{1}{6} N_u (2 N_u + 1) + \frac{3}{2} \right]
\end{aligned}
\end{equation}
\noindent
The numbers of multiplies required per training iteration using BP, OWO-BP, OIG-HWO, and LM are given by
\begin{equation}
\begin{aligned}
M_{bp} = N_v \left[ MN_u + 2N_h(N + 1) + M(N + 6N_h + 4)\right] + N_w
\end{aligned}
\end{equation}

\begin{equation}
\begin{aligned}
% \begin{split}
M_{owo-bp} = N_v \left[ \frac{}{} 2N_h(N + 2) + M(N_u + 1) + M(N + 6N_h + 4) \right.
\left. + \frac{N_u(N_u + 1)}{2} \right] + M_{ols} + N_h(N + 1)
% \end{split}
\end{aligned}
\end{equation}

\begin{equation}
\begin{aligned}
M_{oig} = M_{owo-bp} + N_v[(N+1)(3MN_h + MN  + 2(M+N) + 3) - M(N + 6N_h + 4) - N_h(N+1)] + (N + 1)^3
\end{aligned}
\end{equation}

\begin{equation}
\begin{aligned}
M_{lm} = M_{bp} + N_v[MN_u(N_u + 3N_h(N + 1)) + 4N_h^2 (N+1)^2]  + N_w^3 + N_w^2
\end{aligned}
\end{equation}

\begin{equation}
\begin{aligned}
M_{scg} = 4N_v [ N_h(N+1) + MN_u] + 10[N_h(N+1) + MN_u]
\end{aligned}
\end{equation}
\noindent

Note that $M_{oig}$ consists of $M_{owo-bp}$ plus the required multiplies for calculating optimal input gains. Similarly, $M_{lm}$ consists of $M_{bp}$ plus the required multiplies for calculating and inverting the Hessian matrix. $N^{\delta}_h$ is the number of new hidden units added at each growing step of the cascade correlation algorithm.

\subsection{Approximation Datasets Results}
We take mean square error (MSE) in the approximation datasets as the metric for various algorithm performances. In all data sets, the inputs have been normalized to be zero-mean and unit variance. This way, it becomes clear that OIG’s improved results are not due to a simple, single-data normalization. Table \ref{table:approx_datasets} shows the specifications of the datasets used to evaluate the algorithm performances.

\begin{table*}[h!]
	\centering
	\caption{Specification of approximation datasets}
	\label{table:approx_datasets}
 % \begin{tabular*}{\textwidth}{@{\extracolsep{\fill}}lcccccc@{\extracolsep{\fill}}}
	\begin{tabular}{|p{4cm}|p{2cm}|p{2cm}|p{2cm}|}
		\hline
  \textbf{Datasets}         & \textbf{N}     & \textbf{M} & \textbf{$N_v$}  \\  
  \hline
		Prognostics         & 17             & 9          & 4745  \\ \hline
        Remote Sensing      & 16             & 3          & 5992  \\ \hline
		Federal Reserve     & 15             & 1          & 1049  \\ \hline
		Housing             & 16             & 1          & 22784 \\ \hline
	    Concrete            & 8              & 1          & 1030 \\ \hline
	    White Wine          & 11             & 1          & 4898  \\ \hline
        Parkinson’s         & 16             & 2          & 5875  \\ \hline
	\end{tabular}
\end{table*}

The number of hidden units for each data set is selected by first training a multi-layer perceptron with a large number of hidden units followed by a step-wise pruning with validation \cite{tyagi2020second}. The least useful hidden unit is removed at each step until only one hidden unit is left. The number of hidden units corresponding to the smallest validation error is used for training on that data set. A maximum number of iterations is fixed for each algorithm, along with an early stopping criterion. The maximum training iterations for all algorithms are set to 1000.

For each dataset, we perform a 10-fold training with cross-validation and testing for the proposed OIG-HWO algorithm and compare with those listed in section \ref{sec:Experimental Results}, using the datasets listed in Table \ref{table:approx_datasets}. Two plots are generated for each datasets. The average mean square error (MSE) for training from 10-fold training is plotted versus the number of iterations (shown on a $log_{10}$ scale), and the average training MSE from 10-fold training is plotted versus the cumulative number of multiplies (also shown on a $log_{10}$ scale) for each algorithm. Our results will present the average MSE achieved through training, along with the corresponding computational requirements. Given the constraints on the length of the paper, we have opted to display two plots for the initial dataset, while for the subsequent datasets, we depict the average MSE versus the cumulative number of multiplications. This approach provides a more compelling representation of both the learning and computational aspects of our study.

\subsubsection{Prognostics Dataset}
\label{prognostic_dataset}
\begin{figure}[h!]
\centering
\includegraphics[width=0.9\textwidth,height=100mm]{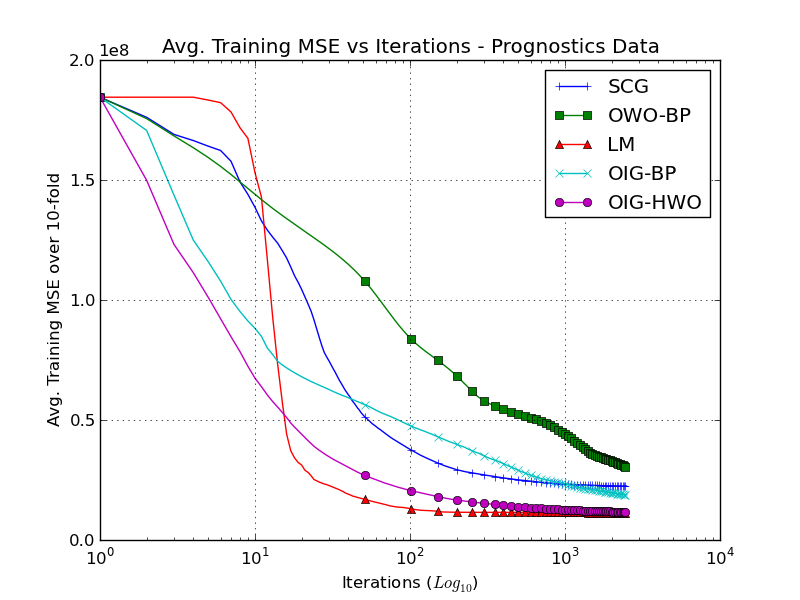}
\caption{Average error vs. Iterations for Prognostics data}
\label{fig:prog_error}
\end{figure}

\begin{figure}[h!]
\centering
\includegraphics[width=0.9\textwidth,height=100mm]{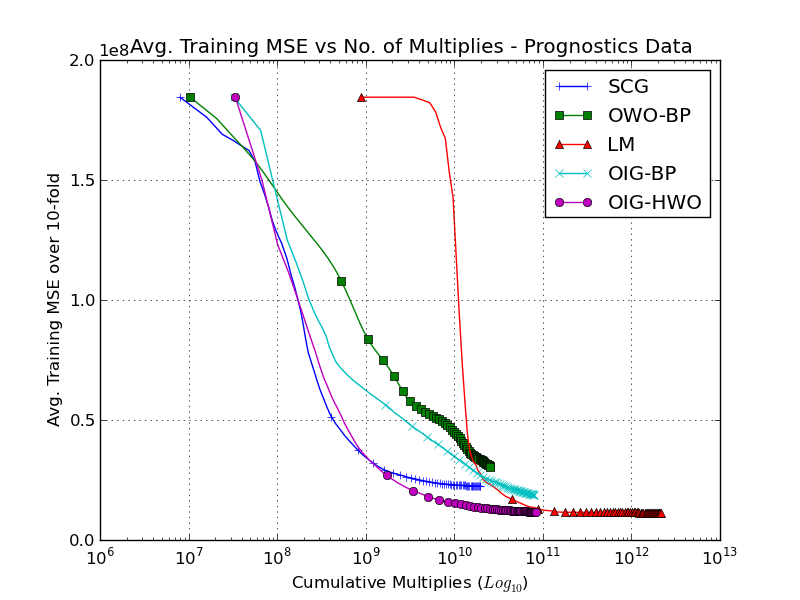}
\caption{Average error vs. cumulative multiplies for Prognostics data}
\label{fig:prognostic_multiplier}
\end{figure}

The Prognostics datasetile, called F-17 \cite{ipnnl_dataset_approx} consists of parameters that are available in the Bell Helicopter health usage monitoring system (HUMS), which performs flight load synthesis, which is a form of prognostics \cite{manry2001signal}. For this data file, 13 hidden units were used for all algorithms. In Figure \ref{fig:prog_error}, the average mean square error (MSE) for training from 10-fold validation is plotted versus the number of iterations for each algorithm. In Figure \ref{fig:prognostic_multiplier}, the average training MSE from 10-fold validation is plotted versus the cumulative number of multiplies. From Figure \ref{fig:prog_error}, the overall training error for the proposed OIG-HWO overlaps with LM, with LM coming out on top by a narrow margin. However, the performance of LM comes with significantly higher computational demand, as shown in Figure \ref{fig:prognostic_multiplier}. Prognostics data is a highly correlated/non-correlated dataset. The proposed OIG-HWO algorithm can give good performance despite these dependent features. The reason is that the input gains checks on the dependent features and reduces their effect while training. This aspect of input gains is not present in other comparing algorithms, and hence they are not as efficiently performing as the proposed algorithm. Another aspect of the features is that distinct distributions are essential in the algorithm's performance. This is proven by evaluating the algorithms using dependent features with and without. From Table \ref{table:Combined best testing MSE for approximation data}, we observe that LM gives the marginally lowest mean squared error followed by OIG-HWO. It is worth emphasizing that the OIG-HWO algorithm achieves significantly lower testing errors than similar algorithms, except LM.

% \begin{figure}[H]
% \centering
% \includegraphics[width=0.8\textwidth,height=50mm]{}
% \caption{Test Test Test}
% \end{figure}

\subsubsection{Remote Sensing Dataset}
\begin{figure}[h!]
\centering
\includegraphics[width=0.9\textwidth,height=100mm]{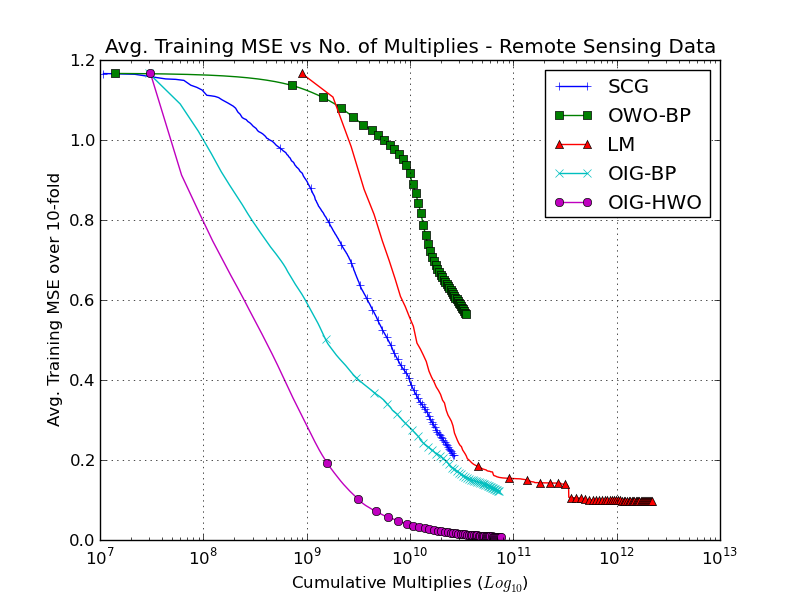}
\caption{Average error vs. cumulative multiplies for Remote Sensing dataset}
\label{fig:remote_sensing_multiplier}
\end{figure}

The Remote Sensing data \cite{ipnnl_dataset_approx} represents the training set for inversion of surface permittivity, the normalized surface rams roughness, and the surface correlation length found in backscattering models from randomly rough dielectric surfaces \cite{fung1992backscattering}. For this dataset, 25 hidden units were used for all algorithms. From Figure \ref{fig:remote_sensing_multiplier}, the average training MSE from the 10-fold procedure for OIG-HWO is better than all other algorithms being compared. Regarding computational cost, the proposed algorithms consume slightly more computation than OWO-BP, SCG, and OIG-BP. However, all algorithms utilize fewer computations about two orders of magnitude than LM. By assigning input gains, the proposed OIG-HWO algorithm computes tailored weights for each input feature, reducing the effect of less important features and extracting useful information from the such high dimensional dataset. From Table \ref{table:Combined best testing MSE for approximation data}, we infer that OIG-HWO performs as the second-best algorithm to LM. However, it is worth noting that LM requires much larger multipliers than OIG-HWO.

\subsubsection{Federal Reserve Dataset}
\begin{figure}[h]
\centering
\includegraphics[width=0.9\textwidth,height=100mm]{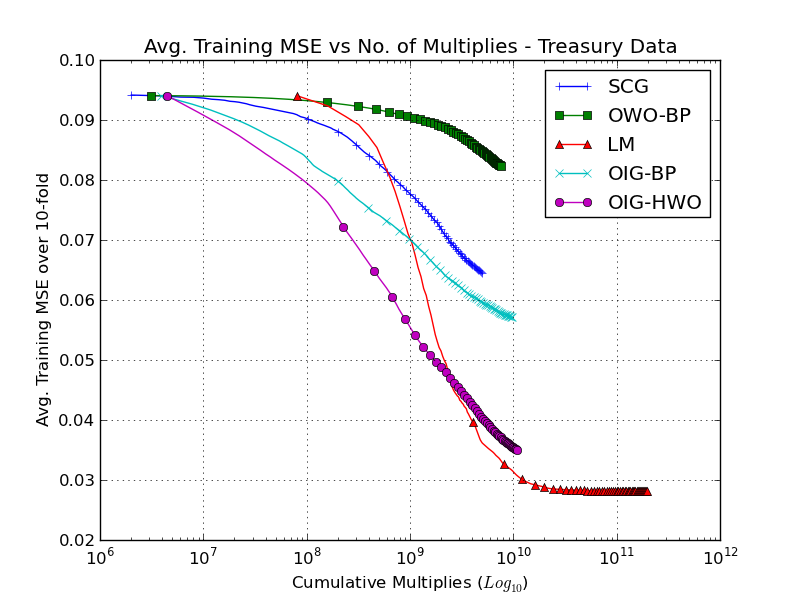}
\caption{Average error vs. cumulative multiplies for Federal Reserve dataset}
\label{fig:federeal_reserve_multiplier}
\end{figure}

The Federal Reserve Economic Data Set \cite{federal_reserve} contains economic data for the USA from 01/04/1980 to 02/04/2000 weekly. From the given features, the goal is to predict the 1-Month CD Rate similar to the US Census Bureau datasets \cite{us_census}. For this data file, called $\textit{TR}$ on its webpage, 34 hidden units were used for all algorithms. Figure \ref{fig:federeal_reserve_multiplier} shows LM had the best overall performance, with OIG-HWO a close. The proposed improvement to OIG performs better than OIG-BP, SCG, and OWO-BP without significant computational overhead. The proposed OIG-HWO algorithm can handle such data as it assigns high weights to features with acceptable variance than to features with very low variance. Results obtained from OIG-HWO act as a testament to the same. From Table \ref{table:Combined best testing MSE for approximation data}, we observe that OIG-HWO performs better than any other comparable algorithms. 

\subsubsection{Housing Dataset}
\begin{figure}[h]
\centering
\includegraphics[width=0.9\textwidth,height=100mm]{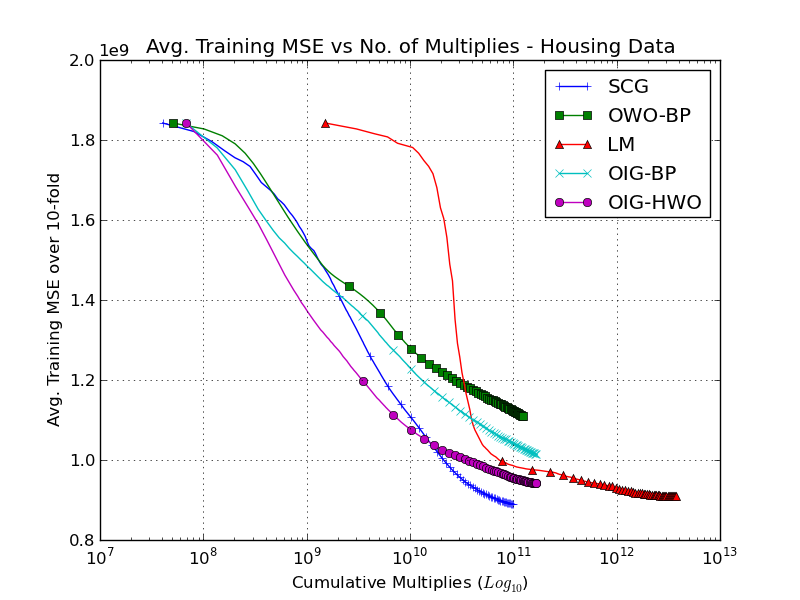}
\caption{Average error vs. cumulative multiplies for Housing Dataset}
\label{fig:housing_data_multiplier}
\end{figure}

The Housing dataset \cite{delve} is designed based on data provided by the US Census Bureau \cite{us_census}. These are all concerned with predicting the median price of houses in a region based on demographic composition and the state of the housing market in the region. House-16H data was used in our simulation, with ‘H’ standing for high difficulty. Tasks with high difficulty have had their attributes chosen to make the modeling more difficult due to higher variance or lower correlation of the inputs to the target. For this dataset, 30 hidden units were used for all algorithms. From Figure \ref{fig:housing_data_multiplier}, the SCG algorithm narrowly came out on top, followed closely by LM and OIG-HWO, respectively. OIG-HWO algorithm adjusts the input gains for both types of distribution and learns to assign lower weights to low variant features. Thus, the columns with almost constant values do not contribute toward the end goal. Table \ref{table:Combined best testing MSE for approximation data} shows the superiority of OIG-HWO over other algorithms. 
 
\subsubsection{Concrete Compressive Strength Dataset}
\begin{figure}[h]
\centering
\includegraphics[width=0.9\textwidth,height=100mm]{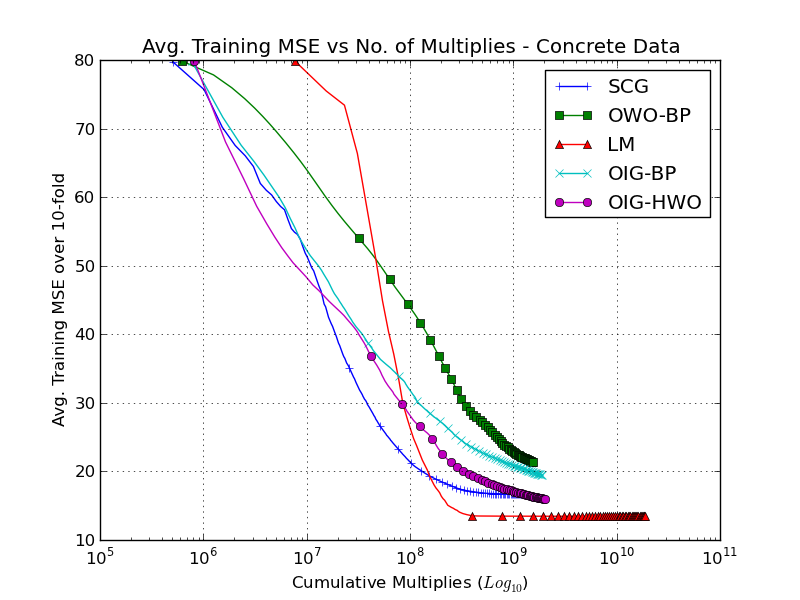}
\caption{Average error vs. cumulative multiplies for Concrete Dataset}
\label{fig:concrete_data_multiplier}
\end{figure}

The Concrete dataset \cite{yeh1998modeling}\cite{concrete} is the actual concrete compressive strength for a given mixture under a specific age (days) determined by laboratory. The concrete compressive strength is a highly nonlinear function of age and ingredients. For this dataset, we trained all algorithms with 13 hidden units. From Figure \ref{fig:concrete_data_multiplier}, LM has the best overall training error, followed by OIG-HWO and closely in third by SCG. Table \ref{table:Combined best testing MSE for approximation data}, also supports the advantage from OIG-HWO over other algorithms. 

\subsubsection{Wine data set}
The Wine dataset \cite{White_Wine} \cite{cortez2009modeling} is related to the wine variant of the Portuguese "Vinho Verde" wine. The inputs include objective tests (e.g., PH values), and the output is based on sensory data (median of at least three evaluations made by wine experts). Each expert graded the wine quality between 0 (very bad) and 10 (very excellent). For this dataset, we trained all algorithms with 24 hidden units. For this data set, Figure \ref{fig:concrete_data_multiplier} shows that LM has the best overall performance, followed very closely by the proposed OIG-HWO. The proposed OIG-HWO algorithm handles dependent features and accounts for the skewness in data, resulting in better results. Table \ref{table:Combined best testing MSE for approximation data} shows that OIG-HWO performs substantially better than the other counterparts. 

\begin{figure}[h]
\centering
\includegraphics[width=0.9\textwidth,height=100mm]{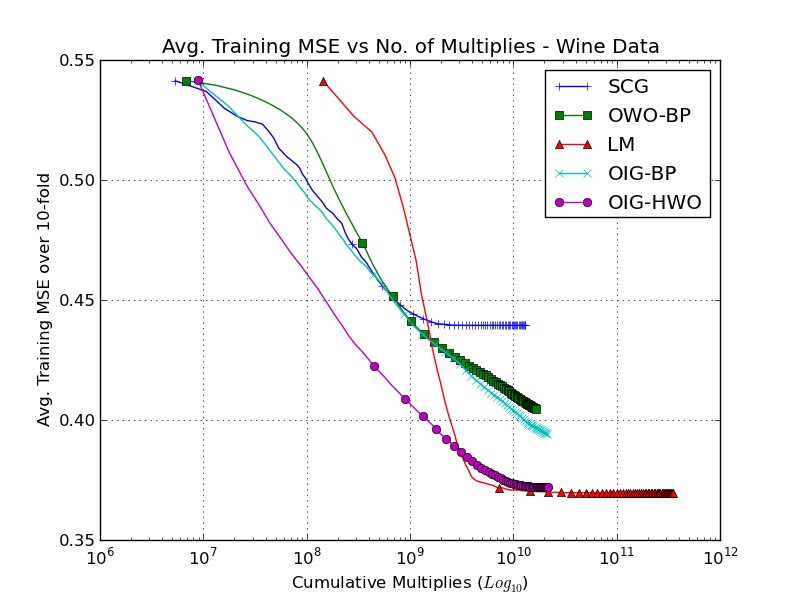}
\caption{Average error vs. cumulative multiplies for White Wine Dataset}
\label{fig:white_wine_multiplier}
\end{figure}

\subsubsection{Parkinson’s Dataset}
\begin{figure}[h]
\centering
\includegraphics[width=0.9\textwidth,height=100mm]{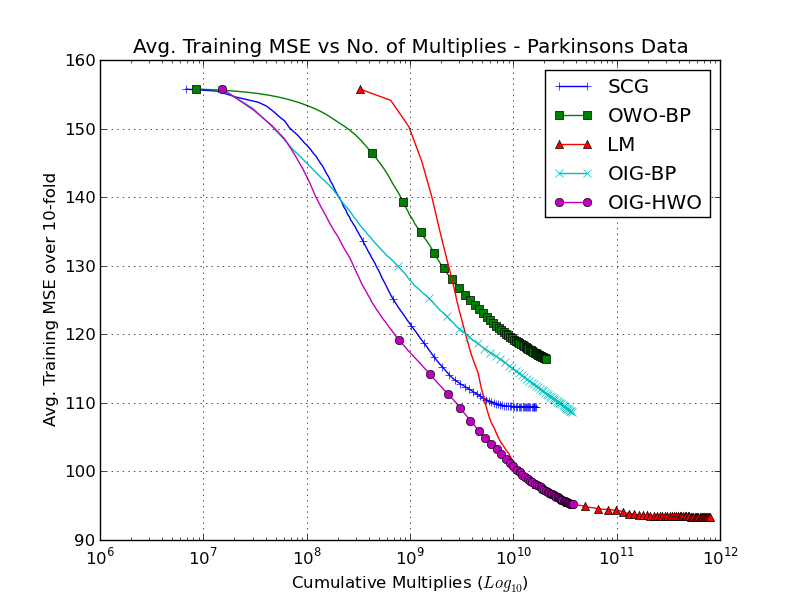}
\caption{Average error vs. cumulative multiplies for Parkinson's Dataset}
\label{fig:parkinson_multiplier}
\end{figure}

The Parkinson's data set \cite{parkinson}, \cite{little2008suitability} comprises a range of biomedical voice measurements from 42 people with early-stage Parkinson's disease recruited to a six-month trial of a telemonitoring device for remote symptom progression monitoring. The main aim of the dataset is to predict the motor and total UPDRS scores ($ 'motor_{UPDRS}' $ and $' total_{UPDRS}' $) from the 16 voice measures. For this data set, LM performed better than the proposed OIG-HWO, followed by the rest. However, LM requires a lot more computations to achieve the slight improvement, as evident in Figure \ref{fig:parkinson_multiplier}. For this dataset, we trained all algorithms with 12 hidden units. From Table \ref{table:Combined best testing MSE for approximation data}, we can observe that OIG-HWO performs better than any other algorithm.

Based on Table \ref{table:Combined best testing MSE for approximation data}, we infer that the proposed OIG-HWO algorithm is consistently much better than other competing algorithms. In addition, OIG-HWO features in the top two performers all the time. The same cannot be said of the other algorithms. Based on Table \ref{table:Combined best testing MSE for approximation data}, it'll be safe to conclude that the suggested improvements to OIG-BP to convert it into OIG-HWO is effective in further reducing the training and validation errors. 

\begin{table*}[h!]
\centering
\caption{10-fold testing MSE results for approximation dataset, (best testing MSE is in bold) } 
\label{table:Combined best testing MSE for approximation data}
% \begin{tabular*}{\textwidth}{@{\extracolsep{\fill}}lcccccc@{\extracolsep{\fill}}}
\begin{tabular}{|p{3cm}|p{2cm}|p{2cm}|p{2cm}|p{2cm}|p{2cm}|}
\hline 
\textbf{Dataset} &  \textbf{OWO-BP}  & \textbf{SCG} & \textbf{LM} & \textbf{OIG-BP} & \textbf{OIG-HWO}\\
\hline
Prognostics      &  3.2752E7        & 2.8224E7     & $\mathbf{1.4093E7}$    & 2.1490E7       & 1.4680E7 \\ \hline
Remote Sensing   &  1.0655          & 0.8627       & 0.2954      & 0.7637         & $\mathbf{0.2094}$ \\\hline
Treasury         &  0.1391          & 0.3245       & 0.1072      & 0.1276         & $\mathbf{0.1036}$ \\\hline
Housing          &  17.2398E8       & 38.0949E8    & 11.9216E8   & 13.2538E8      & $\mathbf{11.7886E8}$ \\\hline
Concrete         &  30.5863         & 73.6012      & 27.7605     & 29.7421        & $\mathbf{27.1604}$ \\ \hline
White Wine       &  0.5222          & 0.5812       & 0.4982      & 0.5027         & $\mathbf{0.4075}$ \\ \hline
Parkinson’s      & 131.1679         & 127.3441     & $\mathbf{123.2571}$    & 124.2698       & 123.57576 \\ \hline
\end{tabular}
\end{table*}

\subsection{Discussion} 
From the training plots and the Table \ref{table:Combined best testing MSE for approximation data}, we deduce the following : 
\begin{enumerate}
    \item OIG-HWO is the top performer in 5 out of the 7 data sets. The following best-performing algorithm is LM by a small margin. However, LM being a second-order method, its performance comes at a significant cost of computation – almost two orders of magnitude greater than the rest. 

    \item In terms of average training error, OWO-BP consistently appears in the last place on 4 of the 7 data sets, while SCG features in the last place on 3 out of 7 data sets. However, being first-order methods, they set the bar for the lowest computation cost. 

    \item Both OIG-BP and OIG-HWO always perform better than their predecessors, OWO-BP and OWO-HWO, respectively, on all data sets, and they are better than SCG. This performance is achieved with minimal computational overhead compared to SCG and OWO-BP, as evident in the error plots vs. number of cumulative multiplications. 

    \item OIG-BP is never better than LM in training and testing MSE and consistently perfrom as the third best for all the datasets. OIG-HWO is always better than OIG-BP. 

    \item LM performs marginally better than OIG-HWO on two data set ($\mathit{Prognostsics}$ and $\mathit{Parkinson's}$ datasets) and has almost identical or worst ten fold testing MSE for the rest of the datasets. 

    \item Ignoring LM, which is a second order computationally heavy, overall, OIG-based algorithms (OIG-BP and OIG-HWO) are consistently in the top two performing algorithms. 
\end{enumerate}

As a general observation, both the OIG-BP and the improved OIG-HWO algorithms consistently outperform the OWO-BP algorithm in all three phases of learning, namely training, validation, and testing. The insertion of OIG into OWO-BP has been found to help enhance its performance. Furthermore, both OIG-BP and the improved OIG-HWO algorithms outperform SCG regarding the average minimum testing error. The OIG-HWO algorithm often performs comparably to LM but with minimal computational overhead. It is worth noting that while OIG-BP is an improvement over OWO-BP, it is not as effective as the OIG-HWO algorithm.

\subsection{Replacement Classifier Datasets}
We compare the OIG-HWO with CG-MLP and SCG using transfer learning. The classification datasets includes MNIST \cite{mnist}, Scrap \cite{scrap}, Fashion-MNIST \cite{scrap}, CIFAR-10 \cite{cifar10}, SVHN \cite{svhn}, Cats-dogs \cite{catsdogs}, Intel Image \cite{Intel_Data}. All the datasets are normalized to zero minimum and maximum one-pixel values. Table \ref{table:replacement_classifier_datasets} shows the specifications of the datasets used to evaluate the algorithm performances. We studied a transfer learning comparison of the OIG-HWO, SCG, and OIG-BP algorithms using normalized classification datasets with pixel values ranging from zero minimum to one maximum. Table \ref{table:replacement_classifier_datasets} outlines the specifications of the datasets used to evaluate algorithm performance.

\begin{table}[h]
	\centering
	\caption{Replacement classifier datasets}
	\label{table:replacement_classifier_datasets}
% \begin{tabular*}{\textwidth}{@{\extracolsep{\fill}}lcccccc@{\extracolsep{\fill}}}
 \begin{tabular}{|p{3cm}|p{2.5cm}|p{2cm}|p{2cm}|p{2cm}|}
		\hline
		Datasets      & N             & M     & $N_v$ & $N_v test$\\ \hline
		MNIST         & 28 x 28 x 1   & 10    & 54000 & 6000   \\ \hline
		Scrap         & 28 x 28 x 1   & 2     & 14382 & 3595  \\ \hline
		Fashion MNIST & 28 x 28 x 1   & 10    & 60000 & 10000    \\ \hline
		CIFAR10       & 32 x 32 x 3   & 10    & 50000 & 10000 \\ \hline
		SVHN          & 32 x 32 x 3   & 10    & 73257 & 26032  \\ \hline
		Cats and Dogs & 32 x 32 x 3   & 2     & 20000 & 5000    \\ \hline
		Intel Image   & 32 x 32 x 3   & 6     & 14034 & 3000    \\ \hline
	\end{tabular}
\end{table}

To create a replacement classifier in a deep learning architecture, we utilized the ResNet-18 architecture \cite{resnet} in the MATLAB 2021 Neural Network toolbox \cite{matlabtoolbox}. We trained ResNet-18 for each dataset, selecting the best validation accuracy ($Pe$) after a certain number of iterations. The training was performed using a learning rate of $1e-4$, $32$ batch size, and Adams \cite{kingma2014adam} optimizer. We found the optimal $N_h$ value by a grid search for various $N_h$ values ($[5,10,15,20,30,100]$)). The feature vector extracted before this final layer was common to all datasets and contained 512 features. The best network was saved, and its final feature layer was extracted as input for each replacement classifier. ResNet-18 requires input images of size 224 x 224 x 3. We implemented an augmented image datastore pipeline with the option $\textit{colorprocessing=gray2rgb}$ to accommodate black and white images. We trained the network using a custom final classification layer tailored to the specific number of classes for each dataset. After training, we replaced the final fully Connected layer of ResNet-18, which has 1000 hidden units, with our replacement classifiers. Table \ref{table:replacement classifier} shows the superiority of OIG-HWO based classifier over other algorithms. 

\begin{table*}[h]
\centering
\caption{10 fold cross validation $Pe$ results for replacement classifier dataset, (best testing $Pe$ is in bold, for optimal $N_h$ values) }
\label{table:replacement classifier}
 \begin{tabular}{|p{3cm}|p{2.5cm}|p{2cm}|p{2cm}|}
\hline
Dataset         &  SCG/Nh       & OIG-BP/Nh    & OIG-HWO/Nh \\
 \hline
MNIST           &  0.39/30      & 0.37/20   & $\mathbf{0.368}$/30 \\\hline
Scrap           & 0.728/20      & 0.554/10  & $\mathbf{0.509}$/30 \\\hline
Fashion MNIST   &  5.705/100    & 5.812/5   & $\mathbf{5.366}$/30 \\ \hline
CIFAR10         & 6.76/100      & 6.599/5   & $\mathbf{6.227}$/100 \\\hline
SVHN            &  3.86025/30   & 3.632/30  & $\mathbf{3.619}$/100 \\ \hline
Cats dogs       &  4.516/100    & 4.504/100 & $\mathbf{4.32}$/10 \\\hline
Intel image     & 9.483/100     & 9.287/10  & $\mathbf{9.02}$/100 \\ \hline
\end{tabular}
\end{table*}

\section{Conclusion and Future Work}
In this study, we investigated the impact of linear input transformations on the training of MLP. To do this, we developed the OIG-HWO algorithm, which optimizes the input gains or coefficients that scale the input data before the MLP processes it. This can improve the performance of the MLP by scaling the input data in an optimal way for the learning process. The OIG-HWO algorithm uses Newton's method to minimize the error function and is capable of handling linearly dependent inputs by "freezing" the weights of the dependent input, effectively removing it from the training process. This can enhance the convergence of the OIG-HWO algorithm. It has been shown that the learning behavior is different for functionally equivalent networks with different input transforms. It has also been shown that learning in the transformed network is equivalent to multiplying the original network's input weight gradient matrix by an autocorrelation matrix. It has also been shown that Newton's algorithm can be used to find the optimal diagonal autocorrelation matrix, resulting in the optimal input gain technique.

Beyond this, the OIG-HWO algorithm has some interesting characteristics desirable for deep learning architectures. Deep learning algorithms have performed exceptionally well in complex applications involving natural language, speech, images, and visual scenes. An underlying issue among these applications is the redundancy in data. Hence a typical pre-processing step in most deep learning applications is to apply Whitening transformation to the raw data. HWO, as mentioned earlier and as shown in Appendix A, is equivalent to back-propagation on Whitened inputs. This means that OIG-HWO could serve as a building block for complex deep-learning architectures that could use the raw data directly without a pre-processing operation.

The OIG technique has been used to substantially speed up the convergence of OWO-BP, which is a two-stage first-order training algorithm. This algorithm was called OIG-BP. The OIG Gauss-Newton Hessian is a weighted average of the input weight Gauss-Newton Hessian, where the weights are elements of the negative input weight gradient matrix. OIG-BP was shown to be sub-optimal in the presence of linearly dependent inputs. Subsequently, OIG was applied to OWO-HWO to create an improved algorithm called OIG-HWO. Results from seven data sets showed that the OIG-based algorithms performed much better than two common first order algorithms with comparable complexity, namely SCG and OWO-BP. They come close to LM regarding the training error, but with orders of magnitude less computation. This is evident in all of the plots of training error versus the required number of multiplies and also from the expressions for the numbers of multiplies. Based on the results, we conclude that OIG-HWO is a strong candidate for shallow learning architectures and performs better than the SCG and OIG-BP algorithms as a replacement classifier.

For future work, the OIG technique can be extended to additional one and two-stage first-order algorithms, including standard BP, to other network types such as RBF networks and additional network parameters, yielding fast second-order methods rival LM's performance but with significantly reduced complexity.

\bibliographystyle{unsrt}
\bibliography{references}

\begin{thebibliography}{10}

\bibitem{stock}
Halbert White.
\newblock Economic prediction using neural networks: The case of ibm daily
  stock returns.
\newblock In {\em ICNN}, volume~2, pages 451--458, 1988.

\bibitem{power_load_forecasting1}
Kang Ke, Sun Hongbin, Zhang Chengkang, and Carl Brown.
\newblock Short-term electrical load forecasting method based on stacked
  auto-encoding and gru neural network.
\newblock {\em Evolutionary Intelligence}, 12:385--394, 2019.

\bibitem{prognostics}
Ahmet Kara.
\newblock A data-driven approach based on deep neural networks for lithium-ion
  battery prognostics.
\newblock {\em Neural Computing and Applications}, 33(20):13525--13538, 2021.

\bibitem{well_logging2}
Yuqing Wang, Qiang Ge, Wenkai Lu, and Xinfei Yan.
\newblock Well-logging constrained seismic inversion based on closed-loop
  convolutional neural network.
\newblock {\em IEEE Transactions on Geoscience and Remote Sensing},
  58(8):5564--5574, 2020.

\bibitem{currency_ER}
Wei Chen, Huilin Xu, Lifen Jia, and Ying Gao.
\newblock Machine learning model for bitcoin exchange rate prediction using
  economic and technology determinants.
\newblock {\em International Journal of Forecasting}, 37(1):28--43, 2021.

\bibitem{LEWIS_robotics}
FL~Lewis, S~Jagannathan, and A~Ye{\c{s}}ildirek.
\newblock Neural network control of robot arms and nonlinear systems.
\newblock In {\em Neural Systems for control}, pages 161--211. Elsevier, 1997.

\bibitem{forecasting}
Karn Meesomsarn, Roungsan Chaisricharoen, Boonruk Chipipop, and Thongchai
  Yooyativong.
\newblock Forecasting the effect of stock repurchase via an artificial neural
  network.
\newblock In {\em 2009 ICCAS-SICE}, pages 2573--2578. IEEE, 2009.

\bibitem{forecasting1}
Bogdan Bochenek and Zbigniew Ustrnul.
\newblock Machine learning in weather prediction and climate
  analyses—applications and perspectives.
\newblock {\em Atmosphere}, 13(2):180, 2022.

\bibitem{speech}
Federico Adolfi, Jeffrey~S Bowers, and David Poeppel.
\newblock Successes and critical failures of neural networks in capturing
  human-like speech recognition.
\newblock {\em Neural Networks}, 162:199--211, 2023.

\bibitem{fingerprint}
Daomiao Wang, Qihan Hu, and Cuiwei Yang.
\newblock Biometric recognition based on scalable end-to-end convolutional
  neural network using photoplethysmography: A comparative study.
\newblock {\em Computers in Biology and Medicine}, 147:105654, 2022.

\bibitem{character_recog}
Xiaoxue Chen, Lianwen Jin, Yuanzhi Zhu, Canjie Luo, and Tianwei Wang.
\newblock Text recognition in the wild: A survey.
\newblock {\em ACM Computing Surveys (CSUR)}, 54(2):1--35, 2021.

\bibitem{face_recog}
Manisha~M Kasar, Debnath Bhattacharyya, and TH~Kim.
\newblock Face recognition using neural network: a review.
\newblock {\em International Journal of Security and Its Applications},
  10(3):81--100, 2016.

\bibitem{tyagi2018automated}
Kanishka Tyagi.
\newblock {\em Automated multistep classifier sizing and training for deep
  learners}.
\newblock PhD thesis, Department of Electrical Engineering, The University of
  Texas at Arlington, Arlington, TX, 2018.

\bibitem{qi2017pointnet}
Charles~R Qi, Hao Su, Kaichun Mo, and Leonidas~J Guibas.
\newblock Pointnet: Deep learning on point sets for 3d classification and
  segmentation.
\newblock In {\em Proceedings of the IEEE conference on computer vision and
  pattern recognition}, pages 652--660, 2017.

\bibitem{duda2012pattern}
Richard~O Duda, Peter~E Hart, and David~G Stork.
\newblock {\em Pattern classification}.
\newblock John Wiley \& Sons, 2012.

\bibitem{wolpert1996lack}
David~H Wolpert.
\newblock The lack of a priori distinctions between learning algorithms.
\newblock {\em Neural computation}, 8(7):1341--1390, 1996.

\bibitem{girosi1989representation}
Federico Girosi and Tomaso Poggio.
\newblock Representation properties of networks: Kolmogorov's theorem is
  irrelevant.
\newblock {\em Neural Computation}, 1(4):465--469, 1989.

\bibitem{Universal_Approx_theorem}
George Cybenko.
\newblock Approximation by superpositions of a sigmoidal function.
\newblock {\em Mathematics of control, signals and systems}, 2(4):303--314,
  1989.

\bibitem{white1990connectionist}
Halbert White.
\newblock Connectionist nonparametric regression: Multilayer feedforward
  networks can learn arbitrary mappings.
\newblock {\em Neural networks}, 3(5):535--549, 1990.

\bibitem{hartman1990layered}
Eric~J Hartman, James~D Keeler, and Jacek~M Kowalski.
\newblock Layered neural networks with gaussian hidden units as universal
  approximations.
\newblock {\em Neural computation}, 2(2):210--215, 1990.

\bibitem{pao1989adaptive}
Yohhan Pao.
\newblock Adaptive pattern recognition and neural networks.
\newblock 1989.

\bibitem{werbos1974beyond}
Paul Werbos.
\newblock {\em Beyond regression: New tools for prediction and analysis in the
  behavioral sciences}.
\newblock PhD thesis, Committee on Applied Mathematics, Harvard University,
  Cambridge, MA, 1974.

\bibitem{GoodBengCour16}
Ian Goodfellow, Yoshua Bengio, and Aaron Courville.
\newblock {\em Deep learning}.
\newblock MIT press, 2016.

\bibitem{Bayes_Classifier}
Bruce~W Suter.
\newblock The multilayer perceptron as an approximation to a bayes optimal
  discriminant function.
\newblock {\em IEEE transactions on neural networks}, 1(4):291, 1990.

\bibitem{geman1992neural}
Stuart Geman, Elie Bienenstock, and Ren{\'e} Doursat.
\newblock Neural networks and the bias/variance dilemma.
\newblock {\em Neural computation}, 4(1):1--58, 1992.

\bibitem{MSE1}
Michael~T Manry, Steven~J Apollo, and Qiang Yu.
\newblock Minimum mean square estimation and neural networks.
\newblock {\em Neurocomputing}, 13(1):59--74, 1996.

\bibitem{MSE2}
Bing-Fei Wu.
\newblock Minimum mean-squared error estimation of stochastic processes by
  mutual entropy.
\newblock {\em International journal of systems science}, 27(12):1391--1402,
  1996.

\bibitem{hecht1992theory}
Robert Hecht-Nielsen.
\newblock Theory of the backpropagation neural network.
\newblock In {\em Neural networks for perception}, pages 65--93. Elsevier,
  1992.

\bibitem{TYAGI20223}
Kanishka Tyagi, Chinmay Rane, and Michael Manry.
\newblock Supervised learning.
\newblock In {\em Artificial Intelligence and Machine Learning for EDGE
  Computing}, pages 3--22. Elsevier, 2022.

\bibitem{CG}
J~Patrick Fitch, Sean~K Lehman, Farid~U Dowla, SHIN-Y Lu, Erik~M Johansson, and
  Dennis~M Goodman.
\newblock Ship wake-detection procedure using conjugate gradient trained
  artificial neural networks.
\newblock {\em IEEE Transactions on Geoscience and Remote Sensing},
  29(5):718--726, 1991.

\bibitem{moller1993scaled}
Martin~Fodslette M{\o}ller.
\newblock A scaled conjugate gradient algorithm for fast supervised learning.
\newblock {\em Neural networks}, 6(4):525--533, 1993.

\bibitem{LM1}
Roberto Battiti.
\newblock First-and second-order methods for learning: between steepest descent
  and newton's method.
\newblock {\em Neural computation}, 4(2):141--166, 1992.

\bibitem{LM2}
Martin~T Hagan and Mohammad~B Menhaj.
\newblock Training feedforward networks with the marquardt algorithm.
\newblock {\em IEEE transactions on Neural Networks}, 5(6):989--993, 1994.

\bibitem{tyagi2021multistage}
Kanishka Tyagi, Chinmay Rane, Bito Irie, and Michael Manry.
\newblock Multistage newton’s approach for training radial basis function
  neural networks.
\newblock {\em SN Computer Science}, 2(5):1--22, 2021.

\bibitem{Tan_2019}
Hong~Hui Tan and King~Hann Lim.
\newblock Review of second-order optimization techniques in artificial neural
  networks backpropagation.
\newblock In {\em IOP conference series: materials science and engineering},
  volume 495, page 012003. IOP Publishing, 2019.

\bibitem{Levenberg1944}
Kenneth Levenberg.
\newblock A method for the solution of certain non-linear problems in least
  squares.
\newblock {\em Quarterly of applied mathematics}, 2(2):164--168, 1944.

\bibitem{Input_means}
Y.~LeCun, L.~Bottou, G.~Orr, and K.~Muller.
\newblock Efficient backprop.
\newblock In G.~Orr and Muller K., editors, {\em Neural Networks: Tricks of the
  trade}, pages 9--50. Springer, 1998.

\bibitem{BP1}
David~E Rumelhart, Geoffrey~E Hinton, and Ronald~J Williams.
\newblock Learning representations by back-propagating errors.
\newblock {\em nature}, 323(6088):533--536, 1986.

\bibitem{Pollack_Kolen}
John Kolen and Jordan Pollack.
\newblock Back propagation is sensitive to initial conditions.
\newblock {\em Advances in neural information processing systems}, 3, 1990.

\bibitem{Hashem1995ImprovingMA}
Sherif Hashem and Bruce Schmeiser.
\newblock Improving model accuracy using optimal linear combinations of trained
  neural networks.
\newblock {\em IEEE Transactions on neural networks}, 6(3):792--794, 1995.

\bibitem{vaswani1601kaiser}
Ashish Vaswani, Noam Shazeer, Niki Parmar, Jakob Uszkoreit, Llion Jones,
  Aidan~N Gomez, {\L}ukasz Kaiser, and Illia Polosukhin.
\newblock Attention is all you need.
\newblock {\em Advances in neural information processing systems}, 30, 2017.

\bibitem{dong2018speech}
Linhao Dong, Shuang Xu, and Bo~Xu.
\newblock Speech-transformer: a no-recurrence sequence-to-sequence model for
  speech recognition.
\newblock In {\em 2018 IEEE international conference on acoustics, speech and
  signal processing (ICASSP)}, pages 5884--5888. IEEE, 2018.

\bibitem{pham2019very}
Ngoc-Quan Pham, Thai-Son Nguyen, Jan Niehues, Markus M{\"u}ller, Sebastian
  St{\"u}ker, and Alexander Waibel.
\newblock Very deep self-attention networks for end-to-end speech recognition.
\newblock {\em arXiv preprint arXiv:1904.13377}, 2019.

\bibitem{dosovitskiy2020image}
Alexander Kolesnikov, Alexey Dosovitskiy, Dirk Weissenborn, Georg Heigold,
  Jakob Uszkoreit, Lucas Beyer, Matthias Minderer, Mostafa Dehghani, Neil
  Houlsby, Sylvain Gelly, Thomas Unterthiner, and Xiaohua Zhai.
\newblock An image is worth 16x16 words: Transformers for image recognition at
  scale.
\newblock 2021.

\bibitem{introducingchatgpt}
Introducing {C}hat {GPT}.
\newblock \url{https://openai.com/blog/chatgpt}, 2023.
\newblock Open {AI}.

\bibitem{tyagi2022radar}
Kanishka Tyagi, Yihang Zhang, John Kirkwood, Shan Zhang, Sanling Song, and
  Narbik Manukian.
\newblock Radar system using a machine-learned model for stationary object
  detection, 2022.
\newblock US Patent App. 17/230,877.

\bibitem{Optimal_Input_Gains}
Sanjeev~S Malalur and Michael Manry.
\newblock Feed-forward network training using optimal input gains.
\newblock In {\em 2009 International joint conference on neural networks},
  pages 1953--1960. IEEE, 2009.

\bibitem{HWO1}
Changhua Yu, Michael~T Manry, and Jiang Li.
\newblock Hidden layer training via hessian matrix information.
\newblock In {\em FLAIRS Conference}, pages 688--694, 2004.

\bibitem{Relu}
Vinod Nair and Geoffrey~E Hinton.
\newblock Rectified linear units improve restricted boltzmann machines.
\newblock In {\em Proceedings of the 27th international conference on machine
  learning (ICML-10)}, pages 807--814, 2010.

\bibitem{gill2019practical}
Philip~E Gill, Walter Murray, and Margaret~H Wright.
\newblock {\em Practical optimization}.
\newblock SIAM, 2019.

\bibitem{biegler1993learning}
Friedrich Biegler-K{\"o}nig and Frank B{\"a}rmann.
\newblock A learning algorithm for multilayered neural networks based on linear
  least squares problems.
\newblock {\em Neural Networks}, 6(1):127--131, 1993.

\bibitem{zhang1999learning}
Zhen Zhang, Weimin Shao, and Hong Zhang.
\newblock A learning algorithm for multilayer perceptron as classifier.
\newblock In {\em IJCNN'99. International Joint Conference on Neural Networks.
  Proceedings (Cat. No. 99CH36339)}, volume~3, pages 1681--1684. IEEE, 1999.

\bibitem{wang1996fast}
Gou-Jen Wang and Chih-Cheng Chen.
\newblock A fast multilayer neural-network training algorithm based on the
  layer-by-layer optimizing procedures.
\newblock {\em IEEE Transactions on Neural Networks}, 7(3):768--775, 1996.

\bibitem{Convex_opt}
Stephen Boyd and Lieven Vandenberghe.
\newblock {\em Convex Optimization}.
\newblock Cambridge University Press, USA, 2004.

\bibitem{Opt_DL_Andrew}
Quoc~V Le, Jiquan Ngiam, Adam Coates, Abhik Lahiri, Bobby Prochnow, and
  Andrew~Y Ng.
\newblock On optimization methods for deep learning.
\newblock In {\em Proceedings of the 28th International Conference on
  International Conference on Machine Learning}, pages 265--272, 2011.

\bibitem{tyagi2014optimal}
Kanishka Tyagi, Nojun Kwak, and Michael~T Manry.
\newblock Optimal conjugate gradient algorithm for generalization of linear
  discriminant analysis based on l1 norm.
\newblock In {\em ICPRAM}, pages 207--212, 2014.

\bibitem{Bishop_PRML}
Christopher~M. Bishop.
\newblock {\em Pattern Recognition and Machine Learning (Information Science
  and Statistics)}.
\newblock Springer-Verlag, Berlin, Heidelberg, 2006.

\bibitem{barton1991matrix}
Simon~A Barton.
\newblock A matrix method for optimizing a neural network.
\newblock {\em Neural Computation}, 3(3):450--459, 1991.

\bibitem{Bayes_discriminant}
Michael Manry, Apollo.~S J, L~S Allen, W~D Lyle, W~Gong, M~S Dawson, and A~K
  Fung.
\newblock Fast training of neural networks for remote sensing.
\newblock {\em Remote Sensing Reviews}, 9:77--96, 1994.

\bibitem{Hinton_ErrorProp}
David~E Rumelhart, Geoffrey~E Hinton, and Ronald~J Williams.
\newblock Learning internal representations by error propagation.
\newblock Technical report, La Jolla Inst for Cognitive Science, University of
  California, San Diego, 1985.

\bibitem{scalero1992fast}
Robert~S Scalero and Nazif Tepedelenlioglu.
\newblock A fast new algorithm for training feedforward neural networks.
\newblock {\em IEEE Transactions on signal processing}, 40(1):202--210, 1992.

\bibitem{raudys2001statistical}
Sarunas Raudys.
\newblock {\em Statistical and Neural Classifiers: An integrated approach to
  design}.
\newblock Springer Science \& Business Media, 2001.

\bibitem{tyagi2020second}
Kanishka Tyagi, Son Nguyen, Rohit Rawat, and Michael Manry.
\newblock Second order training and sizing for the multilayer perceptron.
\newblock {\em Neural Processing Letters}, 51(1):963--991, 2020.

\bibitem{nguyen2016partially}
Son Nguyen, Kanishka Tyagi, Parastoo Kheirkhah, and Michael Manry.
\newblock Partially affine invariant back propagation.
\newblock In {\em 2016 International Joint Conference on Neural Networks
  (IJCNN)}, pages 811--818. IEEE, 2016.

\bibitem{Manry_Effects_of_NonSing_FFN}
Changhua Yu, Michael~T Manry, and Jiang Li.
\newblock Effects of nonsingular preprocessing on feedforward network training.
\newblock {\em International Journal of Pattern Recognition and Artificial
  Intelligence}, 19(02):217--247, 2005.

\bibitem{Manry_Multi_Linear}
Hung-Han Chen, Michael~T Manry, and Hema Chandrasekaran.
\newblock A neural network training algorithm utilizing multiple sets of linear
  equations.
\newblock {\em Neurocomputing}, 25(1-3):55--72, 1999.

\bibitem{ipnnl_dataset_approx}
Regression data files.
\newblock \url{https://ipnnl.uta.edu/training-data-files/regression/}, 2022.
\newblock Image Processing and Neural Networks Lab, The University of Texas
  Arlington.

\bibitem{resnet}
Kaiming He, X.~Zhang, Shaoqing Ren, and Jian Sun.
\newblock Deep residual learning for image recognition.
\newblock {\em IEEE Conference on Computer Vision and Pattern Recognition
  (CVPR)}, pages 770--778, 2016.

\bibitem{manry2001signal}
MT~Manry, H~Chandrasekaran, CH~Hsieh, Yu~Hen Hu, and Jenq-Nenq Hwang.
\newblock Signal processing applications of the multilayer perceptron.
\newblock In {\em Handbook on Neural Network Signal Processing}. CRC Press,
  2001.

\bibitem{fung1992backscattering}
Adrian~K Fung, Zongqian Li, and Kun-Shan Chen.
\newblock Backscattering from a randomly rough dielectric surface.
\newblock {\em IEEE Transactions on Geoscience and remote sensing},
  30(2):356--369, 1992.

\bibitem{federal_reserve}
Function {A}pproximation {R}epo.
\newblock \url{http://funapp.cs.bilkent.edu.tr/DataSets/}, 2011.
\newblock Bilkent University.

\bibitem{us_census}
Us {C}ensus {B}ureau.
\newblock \url{https://www.census.gov/data/datasets.html}, 2022.
\newblock United States Census Bureau.

\bibitem{delve}
Delve {D}atasets.
\newblock \url{http://www.cs.toronto.edu/~delve/data/datasets.html}, 2011.
\newblock The University of Toronto.

\bibitem{yeh1998modeling}
I-C Yeh.
\newblock Modeling of strength of high-performance concrete using artificial
  neural networks.
\newblock {\em Cement and Concrete research}, 28(12):1797--1808, 1998.

\bibitem{concrete}
{UCI} {M}achine {L}earning {R}epository.
\newblock
  \url{https://archive.ics.uci.edu/ml/datasets/concrete+compressive+strength},
  2013.
\newblock University of California, Irvine, School of Information and Computer
  Sciences.

\bibitem{White_Wine}
{UCI} {M}achine {L}earning {R}epository.
\newblock \url{https://archive.ics.uci.edu/ml/datasets/wine+quality}, 2013.
\newblock University of California, Irvine, School of Information and Computer
  Sciences.

\bibitem{cortez2009modeling}
Paulo Cortez, Ant{\'o}nio Cerdeira, Fernando Almeida, Telmo Matos, and Jos{\'e}
  Reis.
\newblock Modeling wine preferences by data mining from physicochemical
  properties.
\newblock {\em Decision support systems}, 47(4):547--553, 2009.

\bibitem{parkinson}
{UCI} {M}achine {L}earning {R}epository.
\newblock \url{https://archive.ics.uci.edu/ml/datasets/parkinsons}, 2013.
\newblock University of California, Irvine, School of Information and Computer
  Sciences.

\bibitem{little2008suitability}
Max Little, Patrick McSharry, Eric Hunter, Jennifer Spielman, and Lorraine
  Ramig.
\newblock Suitability of dysphonia measurements for telemonitoring of
  parkinson’s disease.
\newblock {\em Nature Precedings}, pages 1--1, 2008.

\bibitem{mnist}
Yann LeCun, L{\'e}on Bottou, Yoshua Bengio, and Patrick Haffner.
\newblock Gradient-based learning applied to document recognition.
\newblock {\em Proceedings of the IEEE}, 86(11):2278--2324, 1998.

\bibitem{scrap}
Nalin Kumar, Manuel~Gerardo Garcia, and Kanishka Tyagi.
\newblock Material handling using machine learning system, January~27 2022.
\newblock US Patent App. 17/495,291.

\bibitem{cifar10}
Alex Krizhevsky and Geoffrey Hinton.
\newblock Learning multiple layers of features from tiny images.
\newblock 2009.

\bibitem{svhn}
Yuval Netzer, Tao Wang, Adam Coates, Alessandro Bissacco, Bo~Wu, and Andrew~Y.
  Ng.
\newblock Reading digits in natural images with unsupervised feature learning.
\newblock In {\em NIPS Workshop on Deep Learning and Unsupervised Feature
  Learning 2011}, 2011.

\bibitem{catsdogs}
Omkar~M Parkhi, Andrea Vedaldi, Andrew Zisserman, and C.~V. Jawahar.
\newblock Cats and dogs.
\newblock In {\em 2012 IEEE Conference on Computer Vision and Pattern
  Recognition}, pages 3498--3505, 2012.

\bibitem{Intel_Data}
Intel {I}mage {C}lassification {D}ata.
\newblock
  \url{https://www.kaggle.com/datasets/puneet6060/intel-image-classification},
  2021.
\newblock Intel.

\bibitem{matlabtoolbox}
Matlab {D}eep {L}earning toolbox.
\newblock \url{https://www.mathworks.com/help/deeplearning/ref/resnet18.html},
  2021.
\newblock The MathWorks.

\bibitem{kingma2014adam}
Diederik~P Kingma and Jimmy Ba.
\newblock Adam: A method for stochastic optimization.
\newblock {\em arXiv preprint arXiv:1412.6980}, 2014.

\bibitem{tyagi2019multi}
Kanishka Tyagi and Michael Manry.
\newblock Multi-step training of a generalized linear classifier.
\newblock {\em Neural Processing Letters}, 50:1341--1360, 2019.

\bibitem{golub2012matrix}
Gene~H Golub and Charles~F Van~Loan.
\newblock {\em Matrix computations}, volume~3.
\newblock JHU Press, 2012.

\end{thebibliography}
% \bibliography{cas-refs}

\appendix

\section{Training weights by orthogonal least squares}
OLS is used to solve for the output weights, pruning of hidden units \cite{tyagi2020second}, input units \cite{tyagi2019multi} and deciding on the number of hidden units in a deep learner \cite{tyagi2018automated}. OLS is a transformation of the set of basis vectors into a set of orthogonal basis vectors thereby measuring the individual contribution to the desired output energy from each basis vector.

\par In an autoencoder, we are mapping from an (N+1) dimensional augmented input vector to it's reconstruction in the output layer. The output weight matrix $\mathbf{W_{oh}}$  $\in$ $\Re^{N \times N_{h}}$  and $y_{p}$ in elements wise will be given as 

\begin{equation}  \label{eq:output}
y_p(i) = \sum_{n=1}^{N+1} w_{oh}(i,n) \cdot x_p(n)  
\end{equation}
To solve for the output weights by regression , we minimize the MSE as in \eqref{eq:MSE}. In order to achieve a superior numerical computation, we define the elements of auto correlation $\mathbf{R}$  $\in$ $\Re^{N_{h} \times N_{h}}$   and cross correlation matrix $\mathbf{C}$  $\in$ $\Re^{N_{h} \times M}$   as follows : 

\begin{eqnarray} \label{eq:r_c}
r(n,l)  = \dfrac{1}{N_v}  \sum_{p = 1}^{N_v}O_{p}(n) \cdot O_{p}(l)
~~~~~  c(n,i)  = \dfrac{1}{N_v}  \sum_{p = 1}^{N_v}O_{p}(n) \cdot t_{p}(i)
\end{eqnarray}

Substituting the value of $y_p(i)$  in \eqref{eq:MSE} we get, 
\begin{eqnarray}
E = \frac{1}{N_v}\sum_{p=1}^{N_v}\sum_{m=1}^{M}[t_p(m) -\sum_{k=1} ^{N_h} w_oh(i,k) \cdot O_p(k) ]^2
\end{eqnarray}

Differentiating with respect to $\mathbf{W_{oh}}$  and using \eqref{eq:r_c} we get 

\begin{eqnarray} \label{eq:e_woh}
\dfrac{\partial E}{w_{oh}(m,l)}  = -2 [c(l,m) - \sum_{k=1}^{N_{h} +1} w_{oh}(m,k) r(k,l)]
\end{eqnarray}

Equating \eqref{eq:e_woh} to zero we obtain a  $ M $  set of $N_h + 1 $  linear equations  in $ N_h +1 $ variables. In a compact form it can be written as  

\begin{equation} \label{eq:R_C}
\mathbf{R} \cdot \mathbf{W}^{T} = \mathbf{C}
\end{equation}

By using  orthogonal least square, the solution for computation of weights in \eqref{eq:R_C} will speed up. For convineance, let $N_{u}  = N_{h}+1$ and the basis functions be the hidden units output $ \mathbf{O} $ $\in$ $\Re^{(N_{h}+1) \times 1}$ augmented with a bias of $ \mathbf{1} $. 
For an unordered basis function $ \mathbf{O} $ of dimension $N_u$ , the $m^{th} $ orthonormal basis function $ \mathbf{O^{'}} $ is defines as << add reference >> 

\begin{equation}
O_{m}^{'} = \sum_{k=1}^{m} a_{mk} \cdot O_k 
\end{equation}

Here  $ a_{mk} $ are the elements of triangular matrix  $\mathbf{A} $  $\in$ $\Re^{N_u \times N_u}$

For $m = 1$ 
\begin{equation}
O_{1}^{'} = a_{11} \cdot O_1 
~~~~ a_{11}  = \dfrac{1}{\lVert O \rVert}  = \dfrac{1}{r(1,1)}
\end{equation}

for $ 2 \leq m \leq N_u $, we first obtain 
\begin{equation}
c_i = \sum_{q = 1}^{i} a_{iq} \cdot r(q,m)
\end{equation}

for  $ 1 \leq i \leq m-1 $. Second, we set $ b_m = 1 $ and get 
\begin{equation}
b_{jk} = - \sum_{i = k}^{m=1} c_i \cdot a_{ik} 
\end{equation}

for $ 1 \leq k \leq m-1 $. Lastly we get the coeffeicent $ A_{mk} $ for the triangular matrix $ \mathbf{A} $  as 

\begin{equation}
a_{mk}  = \dfrac{b_k}{[r(m,m)  - \sum_{i =1}^{m-1}c_i ^2]^2}
\end{equation}

Once we have the orthonormal basis functions, the linear mapping weights in the orthonormal system can be found as 

\begin{eqnarray}
w^{'}(i,m) = \sum_{k=1}^{m}a_{mk}c(i,k)
\end{eqnarray}

The orthonormal system's weights  $\mathbf{W^{'}} $ can be mapped back to the original system's weights $\mathbf{W} $ as 

\begin{eqnarray}
w(i,k) = \sum_{m=k}^{N_u}a_{mk} \cdot w_{o}^{'}(i,m)
\end{eqnarray}

In an orthonormal system, the total training error can be written from \eqref{eq:MSE} as 

\begin{eqnarray}
E = \sum_{i=1}^{M}\sum_{p=1}^{N_v}[\langle t_p(i), t_p(i)\rangle  - \sum_{k=1}^{N_u} (w^{'}(i,k))^2]
\end{eqnarray}

Orthogonal least square is equivalent of using the $ \mathbf{QR} $ decomposition \cite{golub2012matrix} and is useful when equation \eqref{eq:R_C} is ill-conditioned meaning that the determinant of $ \mathbf{R} $ is $ \mathbf{0} $.

% \bibliography{sn-bibliography}% common bib file


\end{document}